\documentclass{article}

\usepackage{PRIMEarxiv}

\usepackage[utf8]{inputenc} 
\usepackage[T1]{fontenc}    
\usepackage{hyperref}       
\usepackage{url}            
\usepackage{booktabs}       
\usepackage{amsfonts}       
\usepackage{nicefrac}       
\usepackage{microtype}      
\usepackage{lipsum}
\usepackage{fancyhdr}       
\usepackage{graphicx}       
\graphicspath{{media/}}     
\usepackage{amsmath}
\title{Comparative Sentiment Analysis of Public Perception: Monkey pox vs. COVID-19 Behavioral Insights 
}

\author{
  Author1 \\
  Mostafa Mohaimen Akand Faisal \\
  University of Information Technology and Sciences (UITS) \\
  Gazipur\\
  \texttt{mostafafaisal013@gmail.com} \\
  \And
  Author2 \\
  Rabeya Amin Jhuma \\
  University of Information Technology and Sciences (UITS) \\
  Tongi\\
  \texttt{r.a.jhuma2019@gmail.com} \\
  \And
  Author3 \\
  Jamini Jasim \\
  University of Information Technology and Sciences (UITS) \\
  Tangail\\
}

\begin{document}
\maketitle

\begin{abstract}
The emergence of global health crises, such as COVID-19 and Monkeypox (mpox), has underscored the importance of understanding public sentiment to inform effective public health strategies. This study conducts a comparative sentiment analysis of public perceptions surrounding COVID-19 and mpox by leveraging extensive datasets of 147,475 and 106,638 tweets, respectively. Advanced machine learning models, including Logistic Regression, Naive Bayes, RoBERTa, DistilRoBERTa and XLNet, were applied to perform sentiment classification, with results indicating key trends in public emotion and discourse. The analysis highlights significant differences in public sentiment driven by disease characteristics, media representation, and pandemic fatigue. Through the lens of sentiment polarity and thematic trends, this study offers valuable insights into tailoring public health messaging, mitigating misinformation, and fostering trust during concurrent health crises. The findings contribute to advancing sentiment analysis applications in public health informatics, setting the groundwork for enhanced real-time monitoring and multilingual analysis in future research.
\end{abstract}


\section{Introduction}
The outbreaks of COVID-19 and monkeypox (mpox) have tested global public health responses in distinct ways. Public sentiment during such crises significantly influences trust in authorities, adherence to guidelines, and overall behavioral responses.\\

Social media platforms, particularly Twitter, offer a valuable lens through which to observe these sentiments in real time.\\

This study conducts a comparative sentiment analysis of public perceptions of COVID-19 and mpox. While COVID-19, with its rapid global spread, provoked widespread fear, mpox elicited more varied responses—often shaped by localized outbreaks, transmission differences, and residual pandemic fatigue.

Monkeypox is a viral disease caused by the monkeypox virus (MPXV), belonging to the same family of viruses that causes smallpox, known as the variola virus\cite{b36}.  The World Health Organization (WHO) declared the spread of Monkeypox a global health emergency \cite{b37} due to its sporadic outbreak. \\The department of health and human services secretary of the United States, Xavier Becerra declared this virus a public health emergency on August 4, 2022 \cite{b38}, because of the increased number of cases reported in the US.\\
Monkeypox was first discovered in 1958 when the colonies of monkeys in a research institute in Denmark developed a pox-like disease.\\

The first case in a human was confirmed in 1970 in the Republic of Congo \cite{b39}. Recently, cases have been reported from over 73 countries, and the record shows that the total number of cases reported worldwide as of September 23, 2022, was 65,415, with 24,846 cases from the United States of America \cite{b40}.\\Out of multiple social media platforms, Twitter is highly popular amongst all age groups. As of December 2022, Twitter's audience accounted for over 368 million monthly active users worldwide \cite{b4}.\\Twitter is the most used social media platform amongst journalists \cite{b41} and ranks amongst the most popular social media platforms on a global scale \cite{b4}. The most trusted diagnosis for the virus is the 
polymerase chain reaction (PCR) test, and the available solution remains the development and administration of vaccines \cite{b1}. Studies have shown that Twitter can be an excellent data source for analyzing events worldwide, including health related issues \cite{b42}. For example, since the eruption of COVID 19, social media platforms such as Facebook, Instagram, Pinterest, and Twitter have been the most active means of expressing opinions and sharing information among users \cite{b42}. \\Analyzing content posts is a way to understand the perception of human thought and emotions, as well as reveal the current mood and disposition of the broader human population.\\ Society's reliance on social media for information is enormous, unlike conventional news sources \cite{b2}. The volume of data accessed daily led to the adoption of natural language processing (NLP) f text analytics\cite{b43}. \\Social media reflects trends in society, policy, and public health. Its low cost, accessibility, and broad reach also make it a key tool for business promotion\\
Therefore, social media platforms become a repository for information sources,reviews, and open communication where users' experiences are shared \cite{b2}.  \\

The analysis employs advanced machine learning techniques, including transformer-based models like RoBERTa, DistilRoBER and XLNet, alongside traditional algorithms such as Logistic Regression and Naïve Bayes. By focusing on sentiment polarity and thematic trends, this research aims to provide nuanced insights into how public sentiment evolves during overlapping health crises.\\

The experimental framework involves data collection, preprocessing (including text normalization), labeling, and classification using the ML models.\\ These steps helps to create robust sentiment classification models, enabling a comparison of public sentiment dynamics, enhances understanding of sentiment analysis in public health, providing insights to improve communication, economic impact, and trust, while guiding policymakers through health emergencies.\\
To track sentiment during health crises using machine learning and NLP, with models like Random Forest, SVM, and DistilRoBERTa achieving 85\%-93.48\% accuracy. Challenges include limited multilingual datasets, real-time monitoring, and bot data filtering. Advanced algorithms like BERT/GPT and diverse datasets can improve accuracy and health communication.

\section{Historical Background}
\subsection{Literature Review}

A comprehensive overview of recent research applies sentiment analysis using machine learning to study public responses to monkeypox and COVID-19 across social media and news. These studies reveal key insights into public perceptions and emotional trends. The findings support improved health communication and inform policymaking during crises.\\

The study conducted by Melton et al. (2022) Melton et al. used DistilRoBERTa to analyze COVID-19 vaccine sentiment on Twitter and Reddit, achieving 87\% accuracy. Twitter showed more negative sentiment, while Reddit was more positive. The study highlights the value of sentiment analysis for targeted health communication but notes limitations like lack of real-time tracking and platform bias. Future work may include real-time monitoring, more platforms, and multimodal analysis.  \cite{b1}.

Staphord Bengesi. (2023) also conducted a sentiment analysis on Monkeypox Outbreak utilizeing seven machine learning algorithms: Random Forest (accuracy 90.05\%), Logistic Regression (89.51\%), Support Vector Machine (92.95\%), Multilayer Perceptron (91.86\%), XGBoost (73.46\%), Naïve Bayes (81.09\%), and K-Nearest Neighbors (70.05\%).\\
This study shows that SVM, combined with lemmatization and CountVectorizer, achieved the highest accuracy of 92.95\% in sentiment analysis. It highlights the effectiveness of machine learning in understanding public perceptions during health crises. Future research can benefit from transformer models, multilingual data, and real-time sentiment tracking.  \cite{b2}.

 Similarly, Tareq Al-Ahdal (2022) Tareq Al-Ahdal (2022) analyzed German Twitter data on COVID-19 and monkeypox using LDA and sentiment models like Naïve Bayes, Logistic Regression, and SVM, with SVM achieving 88.3\% accuracy. The study emphasizes the importance of tailored health communication based on public sentiment. Future work could expand to other languages, real-time tracking, and transformer-based models for deeper insights..\cite{b3}
 
Another study was carried out by Nirmalya Thakur  in 2023, analyzying 61,862 tweets using VADER, finding 46.88\% negative, 31.97\% positive, and 21.14\% neutral sentiment on COVID-19 and MPox. This was the first study to analyze both viruses together, identifying key hashtags and topics like "Biden" and "Ukraine." A major limitation was the lack of bot filtering. Future research could explore sentiment changes over time and use broader datasets..\cite{b4}

\begin{table}[htbp]
    \caption{List of related research works}
    \label{tab:related-works}
    \vspace{1em}
    \centering
    \small
    \begin{tabular}{|p{3.5cm}|p{3.8cm}|p{2.8cm}|p{3.5cm}|p{2.5cm}|}
    \hline
    \textbf{Name} & \textbf{Methodology} & \textbf{Performance (Accuracy)} & \textbf{Remarks} & \textbf{Reference} \\ \hline

    Fine-Tuned Sentiment Analysis of COVID-19 Vaccine–Related Social Media Data: Comparative Study &
    Fine-tuned DistilRoBERTa for sentiment classification across social media platforms &
    Overall accuracy: 87\% &
    Identifies platform-specific sentiment differences but lacks real-time analysis and details on model scalability across multiple languages and platforms. &
    Melton et al. (2022) \cite{b1} \\ \hline

    Sentiment Analysis and Text Analysis of the Public Discourse on Twitter about COVID-19 and MPox &
    ML (RF, LR, MLP, SVM, NB, KNN, XGB) &
    RF = 90.05\% \newline LR = 89.51\% \newline MLP = 91.86\% \newline SVM = 92.95\% \newline NB = 81.09\% \newline KNN = 70.05\% \newline XGB = 73.46\% &
    Emphasizes multilingual datasets and effective tokenization but highlights challenges in sarcasm detection and nuanced emotion analysis. &
    Bengesi et al. (2023) \cite{b2} \\ \hline

    Improving Public Health Policy by Comparing the Public Response during the Start of COVID-19 and Monkeypox on Twitter in Germany: A Mixed Methods Study &
    ML (SVM, LR, Naïve Bayes, LDA) &
    SVM = 88.3\% \newline LR = 85.6\% \newline NB = 81.2\% &
    Highlights platform-specific sentiment trends but limited by user demographics and lack of real-time analysis. &
    Al-Ahdal et al. (2022) \cite{b3} \\ \hline

    Sentiment Analysis and Text Analysis of the Public Discourse on Twitter about COVID-19 and MPox &
    ML (VADER, hashtag and word frequency analysis) &
    Negative: 46.88\% \newline Positive: 31.97\% \newline Neutral: 21.14\% &
    First study to compare discourse on both COVID-19 and MPox but lacks filtering of bot-generated tweets, impacting sentiment accuracy. &
    Thakur (2023) \cite{b4} \\ \hline

    \end{tabular}
\end{table}

\pagebreak
The collective research highlights the vital role of sentiment analysis in public health, using machine learning and NLP to track public sentiment during crises. Models like Random Forest, SVM, and DistilRoBERTa achieved high accuracy (85\%–93.48\%), though challenges like limited multilingual data and real-time monitoring remain. Advancing with models like BERT/GPT and broader datasets can enhance accuracy and health communication.

\label{sec:headings}

\section{Research Motivation}
In the wake of global health crises such as COVID-19 and emerging diseases like monkeypox (mpox),Understanding public perception during crises like COVID-19 and mpox is crucial for effective health communication and strategy. The WHO's declaration of mpox as a PHEIC in August 2024 highlights its global significance.\\

WORLD HEALTHORGANIZATION \cite{b6} the Democratic Republic of the Congo (DRC) accounted for 95\% (17,794) of reported cases and 99\% (535) of deaths from mpox in 2024, making it the focal point of response efforts. Additionally, the confirmation of the first mpox cases with clade 1b outside Africa, in Sweden and Pakistan, has heightened global alert levels.The WHO's expert advisory committee has called for a coordinated international response to prevent further spread and save lives. \cite{b5} 

Comparative sentiment analysis of mpox and COVID-19 on Twitter reveals differences in public perceptions, driven by factors like disease characteristics, proximity, and media portrayal. COVID-19 sparked widespread fear, while mpox responses were influenced by stigma and pandemic fatigue. Analyzing Twitter data offers insights to improve public health messaging, combat misinformation, and foster trust in health authorities.\\

This Twitter-based analysis informs targeted communication strategies by revealing how public discourse shapes attitudes toward health crises.\\This research contributes to understanding how public discourse shapes and reflects attitudes toward global health emergencies. The PHEIC declaration for mpox underscores the need for clear and effective communication \cite{b5} , and the study of public sentiment during both mpox and COVID-19 equips policymakers with the knowledge to better navigate these crises and safeguard public health.

\section{Aim and Objectives}
This study aims to compare public perceptions of monkeypox and COVID-19 using Twitter data, analyzing sentiment and thematic trends to understand societal responses during overlapping health crises. The focus is on how perceptions, attitudes, and trust evolve during these pandemics.

\begin{itemize}


    
    \item To systematically analyze and compare public sentiment polarity (positive, negative, neutral) for monkeypox and COVID-19 using advanced sentiment analysis techniques, providing a deeper understanding of public concerns and emotions during these outbreaks.

    \item To identify the key themes, concerns, and misconceptions expressed on Twitter, particularly focusing on how pandemic fatigue, stigma, and media representation influence public discourse and sentiment dynamics.
    
    \item To examine platform-specific behavioral patterns and their implications for public health responses, especially the role of Twitter as a medium for information dissemination and its influence on shaping public perceptions.
    

\end{itemize}
This study uses machine learning to analyze sentiment, incorporating preprocessing to address challenges like social media noise. By comparing monkeypox and COVID-19 sentiment trends, it informs tailored public health messaging and strategies. The research highlights social media's role in capturing real-time public discourse, offering insights for improving communication and public trust during health crises.


\section{Data Analysis}
\subsection{Data Description}

\subsection{COVID-19}
The COVID-19 Twitter dataset, consisting of 147,475 tweets from April to June 2021, captures public discourse during key pandemic phases. It includes detailed metadata, engagement metrics, tweet content, and pre-processed sentiment scores for emotional analysis.

\subsection{Monkeypox}
The Mpox Twitter dataset, with 106,638 tweets from August to September 2022, captures public reactions to the emerging outbreak. It includes metadata, engagement metrics, sentiment scores, and additional fields like hashtags and mentions for trend analysis.\newline



\section{About Dataset}

\textbf{Detailed Role of Each Key Attribute of COVID-19 and Monkeypox :}

\subsubsection{id}
Serves as a unique identifier for tracking tweets, linking sentiment findings to raw data, and ensuring data integrity during analysis

\subsubsection{date/created\_at}
The "created\_at" timestamp allows the study of sentiment changes over time, tracking shifts in response to events, health announcements, or crises, and identifying peak discussion periods.

\subsubsection{original\_text}
Provides raw content for sentiment analysis, extracting emotional tones, opinions, and key topics to inform targeted communication strategies.

\subsubsection{place}
The "place" field provides geographic data, enabling region-specific sentiment analysis and mapping sentiment to locations for understanding localized behavior.

\subsubsection{clean\_tweet}
Contains preprocessed text, removing irrelevant elements to ensure accurate sentiment analysis and improve NLP model performance.

\subsubsection{tweet/text}
The raw content is essential for sentiment analysis, providing the qualitative data needed to extract emotions, opinions, and key topics for shaping targeted communication strategies.

\subsubsection{sentiment}
The "sentiment" label categorizes sentiment as positive, neutral, or negative, enabling analysis, statistical modeling, and visualization of sentiment trends across time or datasets.

\subsubsection{compound, neg, neu, pos}
These sentiment scores quantify the emotional tone of tweets. They are critical for machine-driven sentiment classification and nuanced analysis.

\section{Methodology \& Algorithms}
\subsection{Methodology}
\section*{A. OVERVIEW}
We started the experimental framework, as shown in Figure 1, with data collection. The dataset, consisting of Twitter data, was sourced from Kaggle's repository. We collaboratively carried out the preprocessing steps to clean and prepare the data for subsequent analysis. Our preprocessing stage includes:the Convertation of Text to Lowercase , remove punctuations, remove hashtags, remove mentions, Remove URLs, Remove words with numbers, stopwords, adding tokenization . Emojis were converted into 
text. Preprocessed dataset was normalized via stemming and 
lemmatization. In the data labeling process, we employ two distinct approaches. For the first approach, we apply RoBERTa, a robust transformer-based model, to perform the labeling and in the second approach, we apply TextBlob for sentiment analysis, leveraging its capabilities to determine the sentiment of the text. Lastly, we developed classification models using various machine learning algorithms. In the first approach, we employ two traditional machine learning model Logistic Regression and Naive Bayes, while in the second approach, we apply RoBERTa, DistilRoBERTa, and XLNet."

\begin{figure}[htp]
    \centering
    \includegraphics[width=\textwidth]{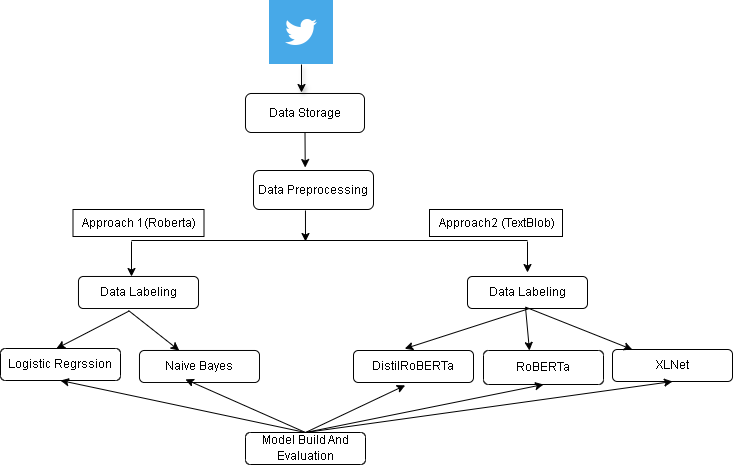}
    \caption{Experimental framework. The figure elucidates a step-by-step methodology for our experiment starting from data collection to 
pre-processing, labeling, and classification algorithms applied with their respective components.}
    \label{fig:wdm}
\end{figure}

\section*{B. DATA COLLECTION}
The data for this analysis was sourced from Kaggle, organized into a CSV file, and includes five features: tweet/text, date/created\_at, place, id, and sentiment.

\section*{C. DATA PREPROCESSING}
Data preprocessing is an integral part of natural language 
processing (NLP) that helps to reconstruct raw text into a 
meaningful format\cite{b7}. In this study, we employed text preprocessing techniques such as lowercase conversion, removal of punctuation, hashtags, mentions, URLs, number-containing words, stop words, tokenization, stemming, and lemmatization. A detailed discussion of each preprocessing task is provided below:

\subsection*{1) Converting Text to Towercase}
Converting text to lowercase is a fundamental preprocessing step in NLP, ensures uniformity by transforming all characters to lowercase, preventing inconsistencies caused by case variations in analysis.\cite{b8} lowercasing as a preprocessing step in text normalization emphasizes in converting all text to lowercase which is an essential technique for handling case variations in text and ensuring that words with different capitalizations are treated as the same.

\subsection*{2) Tokenization}
We employed the Natural Language Toolkit (NLTK) to tokenize the scraped text, a process that involves dividing the text into smaller units, such as individual words. \\
Tokenization is used to break a sentence down into multiple elements, 
called tokens\cite{b9}. IT is a critical step in sentiment analysis, as it facilitates feature extraction and enhances the efficiency of the training process.

\subsection*{3) Stopword Removal}
To facilitate text preprocessing, we utilized the Natural Language Toolkit (NLTK) for tokenization, which involves dividing the scraped text into smaller components. Stopword removal eliminates words with a high frequency of occurrences, such as "which," "and," "in," and "of," while retaining important words. For example, the sentence "The app is good" is changed to "Good app" \cite{b10}.\\ 
Tokenization breaks text into manageable units for analysis, while stopword removal eliminates common, low-value words, improving dataset quality for more accurate sentiment analysis.

\subsection*{4) Removing URLs }
Removing URLs is an essential preprocessing step in NLP to ensure cleaner and more meaningful text data. URLs often do not contribute to the semantic understanding of text and can introduce noise into models. By eliminating 
irrelevant elements, reducing noise and improving the quality of text data for tasks like sentiment analysis and text classification carrying contextual or emotional significance. Most researchers consider that URLs do not carry much information regarding the sentiment of the tweet. Here, Twitter's short URLs are expanded to URLs and are tokenized. Then, the URL matching the tokens are removed from tweets to refine the tweet content\cite{b11}.

\subsection*{5) Removing Mentions}
Removing mentions (@username) in text data reduces noise, ensuring focus on the content and improving the quality and accuracy of sentiment analysis and other NLP tasks.

\subsection*{6) Hashtag, Numeral, and Punctuation Removal}
Hashtags, used for content categorization on social media, are removed from the dataset as they are unnecessary for learning models and do not contribute to the analysis. So we removed Numerals, Repeated words, and 
punctuation using regular expressions (RegExp.) since they do 
not contribute to our data analysis. Doing so helped to fasten 
the learning process and lower memory consumption \cite{b12}. 

\subsection*{7) Emoji and text Conversion }
Emojis, representing emotions and nuances in text, were converted into their corresponding textual format to enhance model training and improve analysis.
The task of predicting emojis has gained paramount significance in the realm of NLP\cite{b13}.

\section*{D. DATASET EXPLORATION }
Before model training, a dataset exploration was conducted to identify frequent words and dominant themes for COVID-19 and Monkeypox. This included word frequency analysis and word cloud generation for insight extraction.\\
\begin{figure}[htp]
    \centering
    \begin{minipage}{0.45\textwidth}
        \centering
        \includegraphics[width=\linewidth]{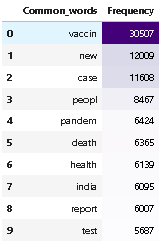}
        \caption{Word Frequency COVID-19}
        \label{fig:wdm1}
    \end{minipage}\hfill
    \begin{minipage}{0.45\textwidth}
        \centering
        \includegraphics[width=\linewidth]{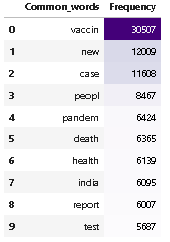}
        \caption{Word Frequency Monkeypox}
        \label{fig:wdm2}
    \end{minipage}
\end{figure}
\subsection{Most Frequent Word COVID-19 :}
The word "vaccine" appears 30,507 times, highlighting its central role in public discussions about COVID-19, particularly around vaccination campaigns, efficacy, and availability.\\

\textbf{Frequently Used Words COVID-19 :}
\begin{itemize}

    \item The word "new" appears 12,009 times, reflecting ongoing developments like emerging variants, guidelines, and cases central to pandemic discussions.
    
    \item "case" (11,608) a pivotal term, emphasizing the public and media focus on the daily or cumulative count of COVID-19 cases during the pandemic.
    
    "report" (6,007) indicates discussions about official updates, news reports, or scientific findings related to COVID-19, reflecting the public's reliance on information sources.

    \item "health" (6,139) 	"Health" reflects an emphasis on public health measures, healthcare systems, and overall well-being during the crisis.
    
    \item The term  "pandem" (6,424)  signifies the global scale of COVID-19 and its significance as a once-in-a-century public health crisis
    
    \item "death" (6,365) highlights discussions about mortality rates, the emotional and societal impact of the pandemic, and the urgency of mitigating its effects.

    \item "Test" (5,687) reflects conversations about COVID-19 testing protocols, availability, accuracy, and the importance of widespread testing to control the pandemic's spread.
    
\end{itemize}

\textbf{Geographical Reference:}
"india" appears 6,095 times, suggesting significant representation of tweets or text related to India in the dataset.\\

\textbf{Public Concerns:}
Words like "peopl" (8,467) indicate discussions or concerns centered on individuals or communities affected by the crises.\\

\subsection{Most Frequent Word Monkeypox :}
The word "vaccine" appears 8,055 times, highlighting the global focus on the development and distribution of vaccines to prevent Monkeypox, reflecting public health campaigns and discussions on vaccine availability and efficacy.\\

\textbf{Frequently Used Words Monkeypox :}
\begin{itemize}

    \item 	The occurrence of "nan" (73,678) indicates missing or null entries in the dataset, suggesting incomplete data for certain topics, time periods, or regions related to Monkeypox, highlighting the need for thorough data cleaning.
    
    \item 	"Case" (8,474) is frequently used in discussions about confirmed Monkeypox infections, case numbers, and tracking the disease's spread, playing a key role in guiding public health responses.
    
    \item The word "de" (6,404) may be a common preposition in languages like French or Spanish, or part of a medical term/abbreviation, requiring clarification to understand its context within the dataset.

    \item The term "health" (6,133) highlights the global focus on public health, healthcare systems, and the response to the Monkeypox outbreak, emphasizing disease prevention and monitoring population health.
    
    \item The frequent appearance of the word "new" (5,614) highlights the evolving nature of the Monkeypox outbreak, including new cases, variants, and guidelines, stressing the need for adaptive public health responses.
    
    \item The term "first" (4,386) likely refers to initial reports or cases of Monkeypox, emphasizing the early stages of the outbreak and public awareness, especially regarding first confirmed cases outside endemic regions.
\end{itemize}

\textbf{Geographical Reference:}
The term "US" (5,453) indicates that a substantial portion of the dataset centers on the United States' response to Monkeypox, likely driven by high media coverage, government actions, and public health measures during the outbreak.

\textbf{Public Concerns:}
The term "People" (5,902) underscores the human-centric aspect of the pandemic, focusing on how individuals and communities are impacted, and is often used in discussions of public health interventions and disease spread.

\subsection{Word Cloud COVID-19}
To complement the insights gained from the word frequency analysis, we utilized word cloud visualization to further explore the dataset. Word clouds provide a visually appealing summary of text, highlighting frequently used terms, which aids in understanding the core content\cite{b28}. They allow for quick comparisons between different texts, enhancing the analytical capabilities of users\cite{b29}. We generated four visualizations: one for the entire dataset and three for sentiment polarities (positive, negative, neutral). Key terms like \textit{vaccine}, \textit{new}, \textit{case}, \textit{people}, \textit{pandemic}, and \textit{death} were prominent, reflecting major public concerns during the COVID-19 pandemic.

\begin{figure}[htp]
    \centering
    \begin{minipage}{0.45\textwidth}
        \centering
        \includegraphics[width=\linewidth]{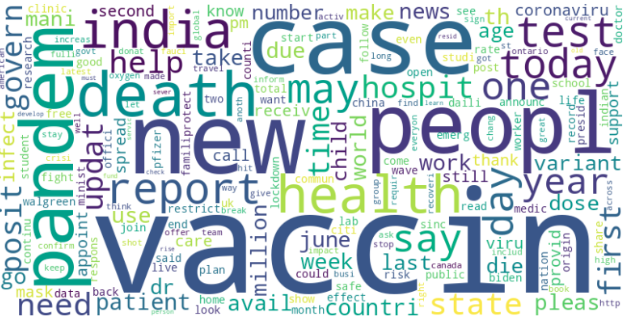}
        \caption{All Sentiment}
        \label{fig:wordcloud_mpox}
    \end{minipage}\hfill
    \begin{minipage}{0.45\textwidth}
        \centering
        \includegraphics[width=\linewidth]{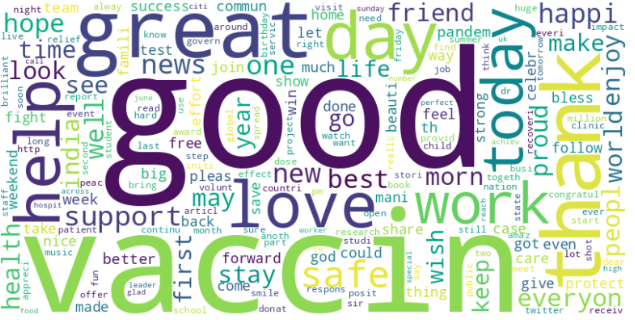}
        \caption{Positive Sentiment}
        \label{fig:wordcloud_mpox}
    \end{minipage}
    
    \vskip\baselineskip
    
    \begin{minipage}{0.45\textwidth}
        \centering
        \includegraphics[width=\linewidth]{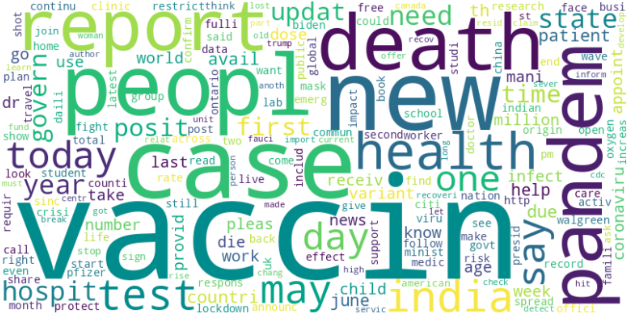}
        \caption{Negative Sentiment}
        \label{fig:wordcloud_mpox}
    \end{minipage}\hfill
    \begin{minipage}{0.45\textwidth}
        \centering
        \includegraphics[width=\linewidth]{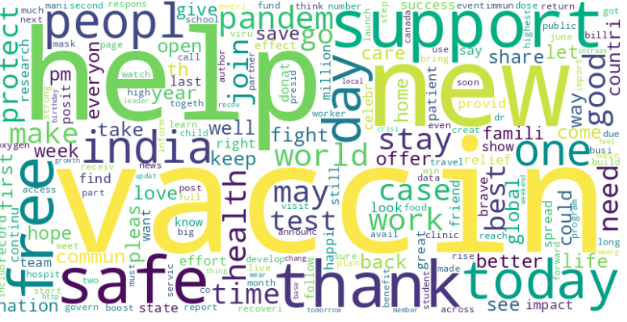}
        \caption{Neutral Sentiment}
        \label{fig:wordcloud_mpox}
    \end{minipage}
\end{figure}
Word Cloud Visualizations of : Display of the most prevalent keywords/common words based on positive, negative, neutral, and the entire dataset sentiment, respectively.

\subsection{WORD CLOUD Monkeypox:}
The most important and frequently appeared key terms in Monkeypox are\textit{vaccine}, \textit{new}, \textit{case}, \textit{de}, \textit{people}, and \textit{first} , highlighting their significance in shaping public sentiment during the Monkeypox pandemic.

\begin{figure}[htp]
    \centering
    \begin{minipage}{0.45\textwidth}
        \centering
        \includegraphics[width=\linewidth]{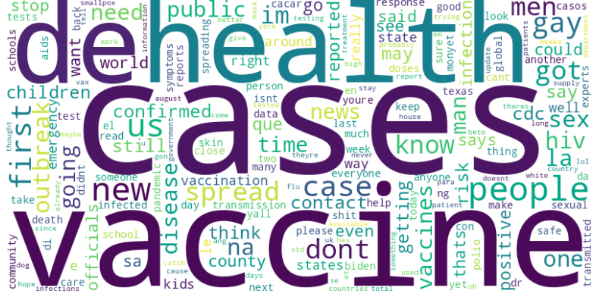}
        \caption{All Sentiment}
        \label{fig:wordcloud_mpox}
    \end{minipage}\hfill
    \begin{minipage}{0.45\textwidth}
        \centering
        \includegraphics[width=\linewidth]{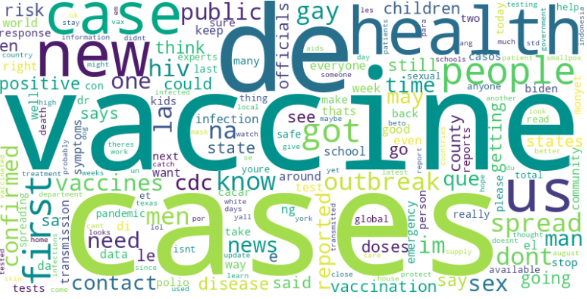}
        \caption{Positive Sentiment}
        \label{fig:wordcloud_mpox}
    \end{minipage}
    
    \vskip\baselineskip
    
    \begin{minipage}{0.45\textwidth}
        \centering
        \includegraphics[width=\linewidth]{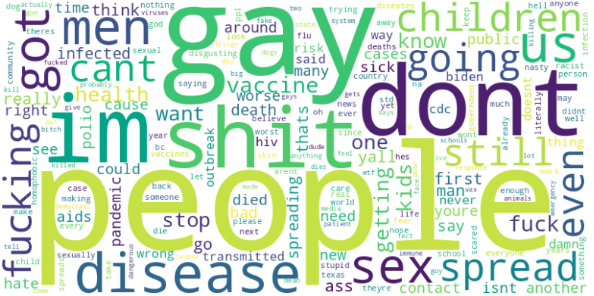}
        \caption{Negative Sentiment}
        \label{fig:wordcloud_mpox}
    \end{minipage}\hfill
    \begin{minipage}{0.45\textwidth}
        \centering
        \includegraphics[width=\linewidth]{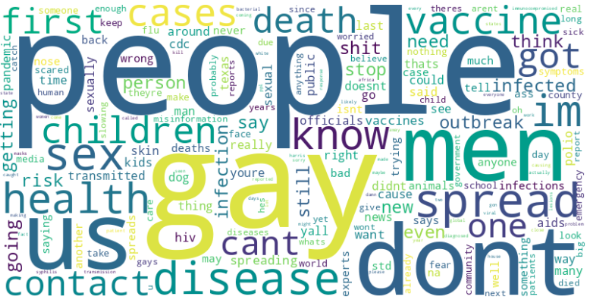}
        \caption{Neutral Sentiment}
        \label{fig:wordcloud_mpox}
    \end{minipage}
\end{figure}
Word Cloud Visualizations of : Display of the most prevalent keywords/common words based on positive, negative, neutral, and the entire dataset sentiment, respectively.

\subsection{Case Fatality Rate (CFR) Comparison}
\begin{figure}[htbp]
    \centering
    \includegraphics[width=0.9\linewidth]{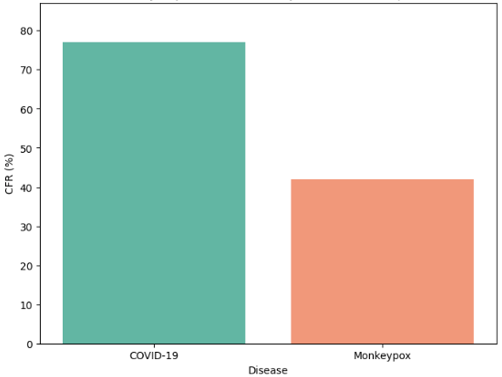}
    \caption{Case Fatality Rate (CFR) Comparison for COVID-19 and Monkeypox.}
    \label{fig:cfr}
\end{figure}

The case fatality rate (CFR), is a measure of the ability of
a pathogen or virus to infect or damage a host in infectious
disease and is described as the proportion of deaths within
a defined population of interest, i.e. the percentage of cases
that result in death\cite{b30}. Figure~\ref{fig:cfr} illustrates the Case Fatality Rate (CFR) for COVID-19 and Monkeypox showing a notable difference between the two. COVID-19 has a CFR of 77.08\%, much higher than Monkeypox's 41.91\%, indicating a greater mortality risk, especially in severe cases of COVID-19.\\
\newline
The Case Fatality Rate (CFR) is calculated using the formula:
\begin{equation}
\text{CFR (\%)} = \left( \frac{\text{Number of Deaths}}{\text{Number of Cases}} \right) \times 100
\end{equation}
\pagebreak

Where:
\begin{itemize}
    \item \textbf{Number of Deaths}: Refers to the total deaths caused by the disease within a defined population and time period.
    \item \textbf{Number of Cases}: Refers to the total confirmed or reported cases of the disease during the same period.
\end{itemize}

\subsection{Age Group Distribution of Cases}

\begin{figure}[htp]
    \centering
    \includegraphics[width=\textwidth]{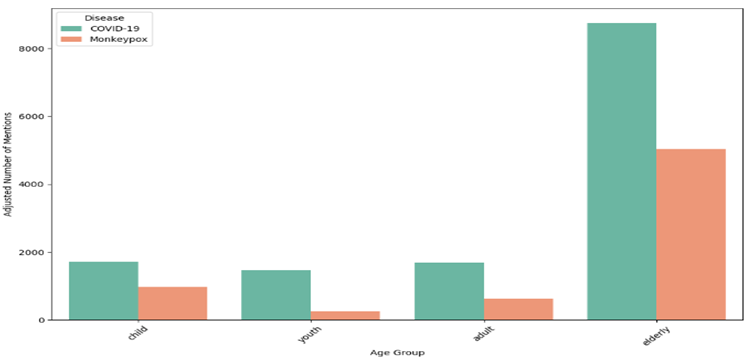}
    \caption{Age Group Distribution of Cases for COVID-19 and Monkeypox.}
    \label{fig:age_distribution}
\end{figure}

Figure~\ref{fig:age_distribution} illustrates the Age Distribution of Cases for COVID-19 and Monkeypox, segmented into four demographic categories: child, youth, adult, and elderly.\\
\pagebreak

The analysis indicates that the elderly population is disproportionately affected by COVID-19, exhibiting significantly higher case numbers compared to Monkeypox. In both diseases, the elderly experience the highest death rates. Studies show that the elderly (over working age) exhibited the highest incidence rates of COVID-19, with a notable increase in cases compared to their population proportion\cite{b31}. However, a notable difference emerges in the second-highest death rate category: for COVID-19, adults hold this position, whereas for Monkeypox, children are the second-most affected group. Similarly, for the third-highest death rate, children are affected in the case of COVID-19, while adults occupy this position for Monkeypox. The proportion of mpox cases among children and adolescents (0-17 years) is approximately 0.46, significantly lower during the ongoing pandemic (0.04) compared to pre-2022 (0.62)\cite{b32}. Approximately 9.7\% of mpox cases reported in the U.S. from May 2022 to May 2023 were among individuals aged over 50 \cite{b33}.  The remaining age groups display a relatively balanced distribution of cases across the two diseases.

\subsection{Gender Distribution of Cases}
\begin{figure}[htp]
    \centering
    \includegraphics[width=10.5cm]{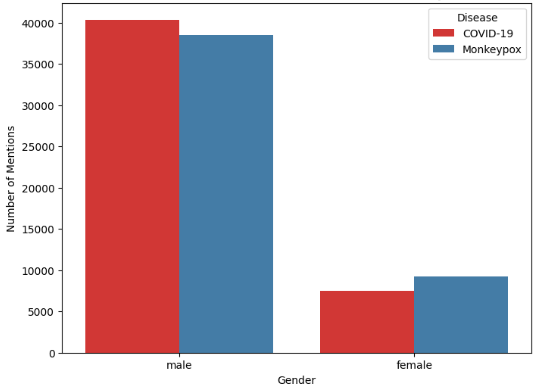}
    \caption{Gender Distribution of Cases for COVID-19 and Monkeypox}
    \label{fig:gender_distribution}
\end{figure}

Figure~\ref{fig:gender_distribution} shows the Gender Distribution of Cases for COVID-19 and Monkeypox, divided into two categories: male and female.\newline

The data shows that male mentions outnumber female mentions for both diseases, with COVID-19 cases slightly higher for males, while female mentions are marginally higher for Monkeypox. Male patients accounted for a higher percentage of COVID-19 cases and exhibited more severe disease manifestations\cite{b34} . In the 2022 outbreak, approximately 91\% of reported cases were men, with only 8.3\% being women\cite{b35}. This trend may reflect differences in reporting, exposure risks, or societal factors, such as occupational exposure, healthcare-seeking behaviors, or biological differences.
\pagebreak

\section*{E. DATA LABELING }
Data labeling assigns meaningful annotations to raw data, enabling machine learning models to interpret and utilize it. In NLP, it involves categorizing text into predefined classes like sentiment or topics, crucial for effective model training and evaluation.\\

In this study, we used two approaches for data labeling. The first approach utilized a pre-trained RoBERTa model for automated sentiment analysis to classify tweets into sentiment categories (positive, negative, and neutral), while also calculating sentiment polarity scores. Domain-specific adjustments were applied, such as amplifying negative sentiments for COVID-19 and emphasizing positive sentiments for Monkeypox, focusing on recovery and vaccination efforts. RoBERTa's contextual understanding ensured high-quality sentiment classification.\\
The second approach involved TextBlob for sentiment analysis, starting with basic sentiment scores and enhancing them for positive keywords (e.g., recovery, vaccine) and lowering them for negative ones (e.g., death, crisis). Sentiments were classified as positive, negative, or neutral, with more emphasis on positive sentiment.

\subsection*{1) RoBERTa}
RoBERTa (Robustly Optimized BERT Approach) is an advanced transformer-based language model developed by Facebook AI, building upon the original BERT architecture. It enhances BERT's performance by training on a larger dataset and modifying key hyperparameters, such as removing the next-sentence prediction objective and employing dynamic masking during training \cite{b13}. RoBERTa can be fine-tuned on labeled data to automate tasks like sentiment analysis by learning to predict labels for new, unseen text.\\

We created a function to load the pre-trained RoBERTa model (\texttt{cardiffnlp/twitter-roberta-base-sentiment}) for classifying tweets into positive, negative, or neutral sentiments. And to adjust domain-specific sentiments, we applied a negative bias for COVID-19 by boosting negative and reducing positive probabilities, and a positive bias for Monkeypox by increasing positive and decreasing negative probabilities.

COVID sentiment polarity is calculated as:\\
             \begin{equation}
    \text{Polarity} = (P_{\text{positive}} - P_{\text{negative}}) \times 3 - 0.5
\end{equation}

\begin{itemize}
    \item Positive and negative scores are weighted to amplify differences.
    \item The offset of $-0.5$ shifts the result slightly toward negative sentiment.
\end{itemize}

This scales the difference between positive and negative sentiment and shifts the result to adjust for the negative bias.

\textbf{Classification}\\
Based on the polarity, tweets are classified as follows:\\
\begin{itemize}
    \item \textbf{Positive:} if \textit{polarity} $> 0.3$.
    \item \textbf{Negative:} if \textit{polarity} $< -0.2$.
    \item \textbf{Neutral:} otherwise.
\end{itemize}

\begin{figure}[htp]
    \centering
    \includegraphics[width=6.5cm]{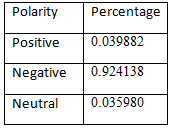}
    \caption{RoBERTa Covid Sentiment Score}
    \label{fig:wdm}
\end{figure}

\begin{figure}[htp]
    \centering
    \includegraphics[width=\textwidth]{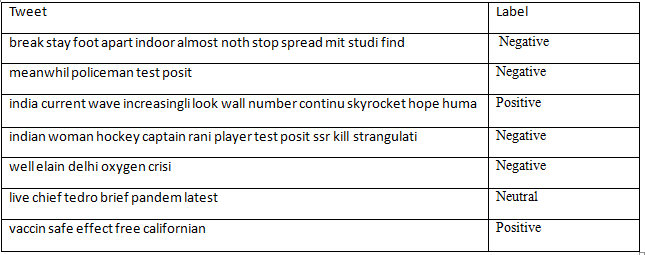}
    \caption{RoBERTa COVID Result.}
    \label{fig:wdm}
\end{figure}

For Monkeypox sentiment snalysis we created a function with positive bias to reflect the hopeful and recovery-oriented nature of Monkeypox-related content.

Monkeypox sentiment polarity is calculated as:\\
              \begin{equation}
    \text{Polarity} = (P_{\text{positive}} - P_{\text{negative}}) \times 4 - 1.2
\end{equation}

\begin{itemize}
    \item A higher weight for positive scores ($\times 4.0$) ensures a stronger influence of positivity on the polarity.
    \item The offset of $+1.2$ shifts the result further toward positive sentiment.
\end{itemize}
This scales the difference between positive and negative sentiment and shifts the result to adjust for the positive bias.\\

\textbf{Classification}\\
Based on the polarity, tweets are classified as:
\begin{itemize}
    \item \textbf{Positive:} if \textit{polarity} $> 0.1$
    \item \textbf{Negative:} if \textit{polarity} $< -0.2$
    \item \textbf{Neutral:} otherwise
\end{itemize}

\begin{figure}[htp]
    \centering
    \includegraphics[width=6.5cm]{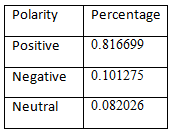}
    \caption{RoBERTa Covid Sentiment Score}
    \label{fig:wdm}
\end{figure}

\begin{figure}[htp]
    \centering
    \includegraphics[width=0.7\textwidth]{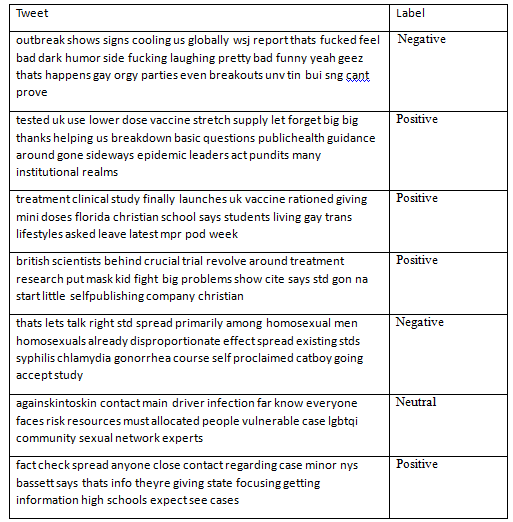} 
    \caption{RoBERTa COVID Result.}
    \label{fig:wdm}
\end{figure}

\subsection*{2) TextBlob}
TextBlob is a Python library for NLP that offers an easy-to-use API for tasks like text preprocessing, sentiment analysis, part-of-speech tagging, and noun phrase extraction, where it computes a polarity score ranging from $-1$ (completely negative) to $+1$ (completely positive). \cite{b15}The algorithms of sentiment analysis mostly focus on defining attitudes, emotions, and even opinions, in a corpus of texts. The TextBlob package is applied to analyze sentiment that requires training words. These texts can be downloaded from NLTK. Moreover, sentiments are considered based on semantic relations and the frequency of use of each text in an input sentence can affect the accuracy of the output as a result.\\
We created a customized function for Monkeypox and COVID-related tweets, applying a negative bias for COVID (amplifying negative sentiment and reducing positive) and a positive bias for Monkeypox (amplifying positive sentiment and reducing negative).\\

COVID Negative sentiment polarity is calculated as:
\[
               \text{polarity} = \min\left(\text{polarity} - (0.6 \times \text{negative\_count}), -0.3\right)   
\]
\textbf{Negative Adjustment:}

\begin{itemize}
    \item For each negative keyword occurrence, the polarity is decreased by $0.6$.
    \item A baseline negative shift of $-0.3$ is applied to handle implicit negative bias.
\end{itemize}

COVID Positive sentiment polarity is calculated as:
\[
             \text{polarity} = \max\left(\text{polarity} + (0.4 \times \text{positive\_count}) + 0.2, 0.2\right)
\]
\textbf{Positive Adjustment:}

\begin{itemize}
    \item For each positive keyword occurrence, the polarity is increased by 0.4.
    \item An extra boost of 0.2 is added if there are no negative keywords.
\end{itemize}

\textbf{Classification}\\
Based on the polarity, tweets are classified as follows:

\begin{itemize}
    \item \textbf{Positive:} Polarity $> 0.15$
    \item \textbf{Negative:} Polarity $< -0.1$
    \item \textbf{Neutral:} Polarity between $-0.1$ and $0.15$
\end{itemize}
\pagebreak

\begin{figure}[htp]
    \centering
    \includegraphics[width=6.5cm]{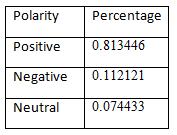}
    \caption{TextBlob COVID Sentiment Score}
    \label{fig:wdm}
\end{figure}
\begin{figure}[htp]
    \centering
    \includegraphics[width=\textwidth]{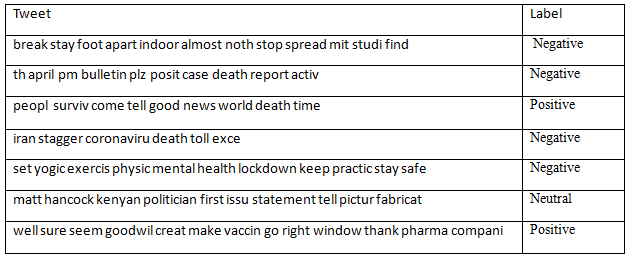}
    \caption{TextBlob COVID Result.}
    \label{fig:wdm}
\end{figure}

For Monkeypox sentiment snalysis we created a function with positive bias to reflect the hopeful and recovery-oriented nature of Monkeypox-related content.\\

Monkeypox positive sentiment polarity is calculated as:
\[
            \text{polarity} = \min\left(\text{polarity} + (0.6 \times \text{positive\_count}) + 0.3, 0.95\right)
\]
\textbf{Positive Adjustment:}

\begin{itemize}
    \item For each positive keyword occurrence, the polarity is increased by $0.6$.
    \item An extra boost of $0.3$ is applied if no negative keywords are found.
    \item The polarity boost is capped at $0.95$ to prevent overly skewed results.
\end{itemize}

Monkeypox Negative sentiment polarity is calculated as:
\[
        \text{polarity} = \max\left(\text{polarity} - (0.2 \times \text{negative\_count}), -0.5\right)
\]
\textbf{Negative  Adjustment:}

\begin{itemize}
    \item For each negative keyword occurrence, the polarity is decreased by $0.2$.
    \item The reduction is capped to ensure that the polarity doesn't drop below $-0.5$.
\end{itemize}

\textbf{Classification}\\
Based on the polarity, tweets are classified as follows:

\begin{itemize}
    \item \textbf{Positive:} If polarity $> 0.05$
    \item \textbf{Negative:} If polarity $< -0.2$
    \item \textbf{Neutral:} Otherwise
\end{itemize}

\begin{figure}[htp]
    \centering
    \includegraphics[width=6.5cm]{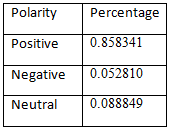}
    \caption{TextBlob Monkeypox Sentiment Score}
    \label{fig:wdm}
\end{figure}
0.85\% were positive 0.05 were negative, and 0.08\% were neutral tweet, shows the frequency of each polarity. shows the frequency of each polarity.

\begin{figure}[htp]
    \centering
    \includegraphics[width=10.5cm]{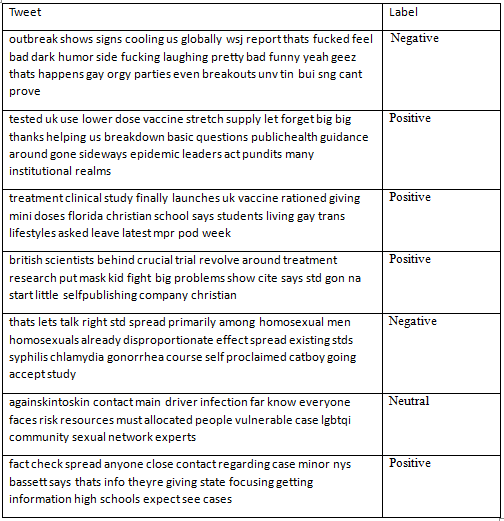}
    \caption{TextBlob Monkeypox Result}
    \label{fig:wdm}
\end{figure}
\newpage
\begin{figure}[htp]
    \centering
    \includegraphics[width=\textwidth]{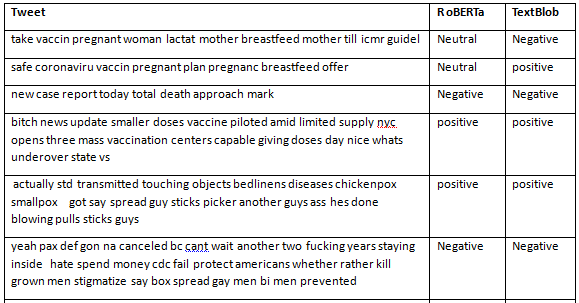}
    \caption{RoBERTa Label vs TextBlob Label disparity.}
    \label{fig:wdm}
\end{figure}

Table 4.16 highlights disparities between the two annotation methods, aligning with previous studies on differences in labeling techniques.Studies have shown that unlike TextBlob, RoBERTa is more focused content \cite{b16}.\newline
In summary, analysis of both techniques revealed that most individuals expressed clear positive or negative opinions on the monkeypox outbreak, indicating a widespread and informed public understanding.

\section*{G. APPLIED ALGORITHMS  }
In this study, we developed and evaluated models using various machine learning algorithms, with 80\% of the data for training and 20\% for testing, assessing performance using accuracy, precision, recall, and F1 score with default sklearn hyperparameters.

\subsection{Logistic Regression}
Logistic regression can be defined as the supervised learning algorithm which predicts the probability of an event occurrence. Its probability is merely based on the selected independent variable vs the dependent variable. This kind of modeling outputs discrete outcome for the given input variable. Logistic regression is mathematically represented in
equation 4.1. \cite{b2}.\\
The probability \( P(Y = 1) \) is defined as:
\begin{equation}
P(Y = 1) = \frac{1}{1 + e^{-(b_0 + b_i X_i)}}
\end{equation}

In Equation (4.2), \( Y \) is the discrete dependent variable (i.e., \( 0, 1, 2, \ldots \)), and \( X \) is an independent variable with subscripts \( i \).

The cost function is defined as:
\begin{equation}
\text{Cost}(h_\theta(x), y) = 
\begin{cases} 
-\log(h_\theta(x)) & \text{if } y = 1 \\ 
-\log(1 - h_\theta(x)) & \text{if } y = 0 
\end{cases}
\end{equation}

 Logistic regression uses binary cross entropy (see Equa
tion 4.3), also known as log loss, as the loss function\cite{b17}

The Binary Cross-Entropy (BCE) is defined as:
\begin{equation}
\text{BCE} = -\frac{1}{N} \sum_{i=0}^{N} \left[ y_i \cdot \log(y_i) + (1 - y_i) \cdot \log(1 - y_i) \right]
\end{equation}

where:
\begin{itemize}
    \item \( N \) is the total number of categories.
    \item \( y \) is the dependent variable.
\end{itemize}

Logistic Regression is a robust, efficient, and interpretable choice for multiclass NLP tasks, especially when paired with effective feature engineering methods like TF-IDF or word embeddings. It serves as an excellent starting point for text classification tasks.\\

\subsection{Naive Bayes}

The Naive Bayes algorithm is a fundamental probabilistic classifier widely used in natural language processing (NLP) due to its simplicity and efficiency. It operates on the principle of feature independence, assuming that the presence of a particular feature in a class is independent of the presence of any other feature. This assumption, while "naive," allows for rapid computation and effective performance in various NLP tasks such as text classification, spam detection, and sentiment analysis \cite{b24}.

\subsection*{1)Text Preprocessing}

Before applying the model, text data is preprocessed (e.g., lowercased, tokenized, and cleaned of stopwords, punctuation, and unnecessary characters).

\begin{itemize}
    \item \textbf{Input (\( X \))}: Cleaned textual data.
    \item \textbf{Labels (\( y \))}: Sentiment labels (e.g., positive, negative, or neutral).
\end{itemize}

\subsection*{Train-Test Split}

The dataset is divided into training and testing subsets:
\[
X = \text{processed\_text\_covid}, \quad y = \text{covid\_sentiment}
\]
\[
X_{\text{train}}, X_{\text{test}}, y_{\text{train}}, y_{\text{test}} = \text{train\_test\_split}(X, y, \text{test\_size}=0.2, \text{random\_state}=42)
\]
\textbf{Purpose:} The training set is used to fit the model, and the test set evaluates its performance.  
\textbf{Test size:} 20\% of the data is reserved for testing.

\subsection*{TF-IDF Vectorization}

The TF-IDF (Term Frequency-Inverse Document Frequency) method converts text into a numerical matrix. It emphasizes important terms by reducing the weight of commonly occurring words across documents.

\subsection*{Term Frequency (TF)}
The term frequency (\( \text{TF} \)) is the frequency of term \( t \) in document \( d \):
\begin{equation}
\text{TF}(t,d) = \frac{f(t,d)}{N_d}
\end{equation}

\begin{equation}
\text{Polarity} = (P_{\text{positive}} - P_{\text{negative}}) \times 4 - 1.2
\end{equation}

Where:
\begin{itemize}
    \item \( f(t,d) \): Frequency of term \( t \) in document \( d \).
    \item \( N_d \): Total number of terms in document \( d \).
\end{itemize}

\subsection*{Inverse Document Frequency (IDF)}
The inverse document frequency (\( \text{IDF} \)) measures the importance of a term across the entire dataset:
\begin{equation}
\text{IDF}(t) = \log\left( \frac{N}{1 + n_t} \right)
\end{equation}

Where:
\begin{itemize}
    \item \( N \): Total number of documents.
    \item \( n_t \): Number of documents containing the term \( t \).
\end{itemize}
Logarithmic smoothing is applied to prevent division by zero.

\subsection*{TF-IDF Score}
The TF-IDF score is the product of term frequency and inverse document frequency:
\begin{equation}
\text{TF-IDF}(t,d) = \text{TF}(t,d) \cdot \text{IDF}(t)
\end{equation}

\textbf{max\_features = 5000:} Limits the vectorizer to the top 5000 terms based on TF-IDF scores.

\subsection*{Multinomial Naive Bayes Classifier}

The Naive Bayes algorithm assumes that features are conditionally independent given the class.

\subsection*{Bayes' Theorem}
Bayes' Theorem is used to calculate the posterior probability of a class \( C \) given input \( X \):
\begin{equation}
P(C \mid X) = \frac{P(X \mid C) \cdot P(C)}{P(X)}
\end{equation}

Where:
\begin{itemize}
    \item \( P(C \mid X) \): Posterior probability of class \( C \) given input \( X \).
    \item \( P(X \mid C) \): Likelihood of data \( X \) given class \( C \).
    \item \( P(C) \): Prior probability of class \( C \).
    \item \( P(X) \): Evidence or marginal likelihood of data \( X \).
\end{itemize}

\subsection*{Multinomial Naive Bayes}
For text data, \( X \) is a vector of term counts or frequencies, and \( P(X \mid C) \) is modeled as:
\begin{equation}
P(X \mid C) = \prod_{i=1}^{n} P(x_i \mid C)
\label{eq:product_rule}
\end{equation}

Where:
\begin{itemize}
    \item \( x_i \): Individual feature (word or term).
    \item \( P(x_i \mid C) \): Probability of \( x_i \) occurring in class \( C \).
\end{itemize}

\subsection*{Class Prediction}
The classifier predicts the class \( C_{\max} \) with the highest posterior probability:

\begin{equation}
C_{\max} = \arg\max_{C \in C} \left[ \log P(C) + \sum_{i=1}^{n} \log P(x_i \mid C) \right]
\end{equation}

Logarithmic probabilities are used for numerical stability.

Naive Bayes is a simple, efficient, and interpretable choice for multiclass NLP tasks, particularly well-suited for text classification when paired with effective feature extraction methods like TF-IDF. Its probabilistic foundation allows for fast computation and scalability, making it an excellent baseline model for handling large-scale text datasets.\\

\subsection{RoBERTa}

RoBERTa (Robustly optimized BERT approach) \cite{b18} is developed by improving BERT and therefore share many similar configurations. We can observe from the GLUE leader-board \cite{b20} that RoBERTa performs better than BERT. RoBERTa improves upon BERT by using more training data, dynamic masking, longer sequences, and removing next sentence prediction, effectively tuning BERT by scaling data and hyperparameters. RoBERTa has the same architecture as BERT while it is pre-trained on a large data corpus \cite{b21}. RoBERTa is a state-of-the-art text classification model known for its strong contextual understanding and exceptional performance across various NLP tasks.

\subsection*{1) Text Input Representation}
RoBERTa is based on the Transformer architecture, which processes input text as sequences of tokens. Here's how text is prepared:

\subsection*{2)Input Text Preparation}
The raw input text is split into subword tokens using Byte Pair Encoding (BPE). For example:

\begin{itemize}
    \item Input text: \textit{"Natural Language Processing"}
    \item Tokenized text: \texttt{["Natural", "Language", "Pro", "cess", "ing"]}
\end{itemize}

Special tokens are added:
\begin{itemize}
    \item \texttt{[CLS]}: Classification token, which acts as a summary representation for the entire sequence.
    \item \texttt{[SEP]}: Separator token, marking the end of a sentence or segment (useful in tasks with multiple sentences).
\end{itemize}

The final input to the model looks like this:
\[
\text{Input} = [\texttt{[CLS]}, \text{Token}_1, \text{Token}_2, \ldots, \text{Token}_n, \texttt{[SEP]}]
\]

\subsection*{3)Embedding Layer}
Each token is converted into a dense numerical vector using the embedding layer. These embeddings capture semantic and positional information about the tokens:

\[
\text{Embedding}(X) = \text{Token Embeddings} + \text{Positional Embeddings}
\]

\begin{itemize}
    \item \textbf{Token Embeddings}: Represent the meaning of each word/subword.
    \item \textbf{Positional Embeddings}: Capture the order of tokens in the sequence.
\end{itemize}

For example, the token \textit{"Natural"} might be mapped to a dense vector: \([0.45, 0.13, \ldots]\). Positional embeddings ensure that \textit{"Natural"} is recognized as the first word in the sequence.

\subsection*{4)Transformer Layers (Contextual Encoding)}
The Transformer is the heart of RoBERTa. It captures contextual relationships between tokens using self-attention and feed-forward layers.

\subsection*{5)Self-Attention Mechanism}
Self-attention computes a weighted representation of each token based on its relationship to all other tokens in the sequence:

\begin{equation}
\text{Attention}(Q, K, V) = \text{softmax}\left(\frac{QK^\top}{\sqrt{d_k}}\right)V
\end{equation}

\begin{itemize}
    \item \(Q, K, V\): Query, Key, and Value matrices derived from the token embeddings.
    \item \(d_k\): Dimensionality of the key vectors.
\end{itemize}

Self-attention allows the model to focus on relevant words for a given token. For example, in the sentence \textit{"RoBERTa is a transformer model"}, the token \textit{"transformer"} may attend strongly to \textit{"RoBERTa"} for context.

\subsection*{6)Output of Transformer}
After passing through multiple transformer layers, each token has a contextualized embedding that incorporates the meaning of the token in the context of the entire sentence. For the \texttt{[CLS]} token:

\[
h_{\texttt{[CLS]}} = \text{Output of Transformer Layers for \texttt{[CLS]}}
\]

\subsection*{7)Classification Head}
The final layer of RoBERTa is a classification head, which maps the \texttt{[CLS]} token's embedding to class probabilities.

\subsection*{8)Linear Transformation}
The \texttt{[CLS]} token's contextualized embedding \(h_{\texttt{[CLS]}}\) is passed through a fully connected (dense) layer:

\begin{equation}
z = W \cdot h_{\texttt{[CLS]}} + b
\end{equation}

\begin{itemize}
    \item \(W\): Weight matrix.
    \item \(b\): Bias vector.
    \item \(z\): Logits (raw scores) for each class.
\end{itemize}

\subsection*{9)Softmax Activation}
The logits are converted into probabilities using the softmax function:

\begin{equation}
P(y_i \mid X) = \frac{\exp(z_i)}{\sum_{j=1}^C \exp(z_j)}
\end{equation}

\begin{itemize}
    \item \(P(y_i \mid X)\): Probability of class \(i\) given the input \(X\).
    \item \(C\): Number of classes.
\end{itemize}

The model predicts the class with the highest probability.

\section*{10)Training Objective}
RoBERTa is fine-tuned on text classification tasks using the cross-entropy loss:
\begin{equation}
L = -\sum_{i=1}^C y_i \log\left(P(y_i \mid X)\right)
\end{equation}

\begin{itemize}
    \item \(y_i\): Ground truth label for class \(i\) (one-hot encoded).
    \item \(P(y_i \mid X)\): Predicted probability for class \(i\).
\end{itemize}

The model minimizes this loss during training to improve its predictions.

\subsection{XLNet}

XLNet is a state-of-the-art model for text classification that outperforms BERT by using a permutation-based training objective to capture bidirectional context while retaining autoregressive properties. XLNet's architecture enables it to consider the entire context of a word, which is crucial for disambiguating meanings of polysemous words. This is supported by findings that contextual embeddings can reflect nuanced meanings based on surrounding words \cite{b22}.

\subsection*{1) Input Text Representation}
XLNet is based on the Transformer architecture, which represents text as sequences of tokens. The input preparation includes tokenization and special token insertion.

\subsection*{2) Input Text Preparation}
The raw input text is tokenized into subword units using the XLNet tokenizer. Special tokens are added to mark sequence boundaries:
\begin{itemize}
    \item \textbf{[CLS]}: Represents the entire sequence for classification tasks.
    \item \textbf{[SEP]}: Marks the end of a sequence or segment.
\end{itemize}

\textbf{Example:}
\begin{itemize}
    \item Input text: \textit{``Monkeypox is a global health concern.''}
    \item Tokenized text: [``Monkey'', ``pox'', ``is'', ``a'', ``global'', ``health'', ``concern'', ``.'']
    \item Final input sequence: [CLS], Monkey, pox, is, a, global, health, concern, ., [SEP]
\end{itemize}

\subsection*{3) Embedding Layer}
Each token is mapped to a dense vector in the embedding layer, combining:
\begin{itemize}
    \item \textbf{Token Embeddings}: Capture the semantic meaning of tokens.
    \item \textbf{Positional Embeddings}: Encode the order of tokens in the sequence.
\end{itemize}

For example, the token ``Monkey'' might map to $[0.12, 0.87, \dots]$. Positional embeddings ensure that the order of tokens, such as ``global'' preceding ``health'', is preserved.

\subsection*{4) Transformer Layers (Contextual Encoding)}
The Transformer layers form the core of XLNet. These layers use self-attention mechanisms to model contextual relationships between tokens.

\subsection*{5) Permutation-Based Training}
XLNet uses a permutation-based language modeling objective. Instead of predicting masked tokens, it predicts tokens based on all possible permutations of the input sequence, capturing bidirectional context.

\subsection*{6) Self-Attention Mechanism}
Self-attention computes the relationship between tokens using:
\begin{equation}
\text{Attention}(Q, K, V) = \text{softmax}\left(\frac{Q K^\top}{\sqrt{d_k}}\right)V
\end{equation}

Where:
\begin{itemize}
    \item $Q, K, V$: Query, Key, and Value matrices derived from token embeddings.
    \item $d_k$: Dimensionality of the key vectors.
\end{itemize}

\subsection*{7) Classification Head}
After contextual encoding, the [CLS] token embedding is passed to a classification head to predict sentiment labels.

\subsection*{8) Linear Transformation}
The contextualized [CLS] embedding, $h_{\text{[CLS]}}$, is transformed into logits using:
\begin{equation}
z = W \cdot h_{\text{[CLS]}} + b
\end{equation}

Where:
\begin{itemize}
    \item $W$: Weight matrix.
    \item $b$: Bias vector.
    \item $z$: Logits (raw scores) for sentiment classes.
\end{itemize}

\subsection*{9) Softmax Activation}
The logits are converted to class probabilities:
\begin{equation}
P(y_i|X) = \frac{\exp(z_i)}{\sum_{j=1}^C \exp(z_j)}
\end{equation}

Where:
\begin{itemize}
    \item $P(y_i|X)$: Probability of class $i$ given input $X$.
    \item $C$: Number of sentiment classes.
\end{itemize}

The softmax cross-entropy loss is commonly used, combining the softmax activation with cross-entropy to measure the difference between predicted and actual distributions \cite{b23} .

\subsection*{10) Training Objective}
The model is fine-tuned for sentiment analysis using the cross-entropy loss function:
\begin{equation}
L = -\sum_{i=1}^C y_i \log P(y_i|X)
\end{equation}

Where:
\begin{itemize}
    \item $y_i$: Ground truth label (one-hot encoded).
    \item $P(y_i|X)$: Predicted probability for class $i$.
\end{itemize}

\subsection*{11) Model Training and Evaluation}
The XLNet model was trained for 5 epochs using the Adam optimizer with a learning rate of $2 \times 10^{-5}$. The loss and accuracy were monitored on a validation set. After training, the model's performance was evaluated using metrics such as precision, recall, and F1 score.

XLNet excels in sentiment analysis through its permutation-based training and rich contextual encoding, making it ideal for tasks demanding nuanced language understanding and accurate classification.

\subsection{DistilRoBERTa}
DistilRoBERTa is a lightweight version of RoBERTa that retains 97\% of its language understanding capabilities while being 60\% faster. It is widely used for tasks like sentiment analysis due to its efficiency and robustness. DistilRoBERTa is a distilled version of the RoBERTa model, designed for natural language processing tasks \cite{b23}.

\subsection*{1) Input Text Representation}

The labels (e.g., positive, negative, neutral) are categorical strings. Machine learning models require numerical inputs, so we transform them into integers using \texttt{LabelEncoder}. The transformation can be mathematically expressed as:

\begin{equation}
    y_{\text{encoded}} = f(y_{\text{original}})
\end{equation}

Where:
\begin{itemize}
    \item \( y_{\text{original}} \): Original categorical labels.
    \item \( f \): Label encoding function mapping categories to integers.
\end{itemize}

For example:
\[
\text{positive} \rightarrow 0, \quad \text{negative} \rightarrow 1, \quad \text{neutral} \rightarrow 2
\]

\subsection*{2) Splitting the Dataset}

We split the dataset into training and testing subsets using the formula:
\begin{equation}
D = D_{\text{train}} \cup D_{\text{test}}
\end{equation}

Where:
\begin{itemize}
    \item \( D_{\text{train}} \): Subset used for training the model.
    \item \( D_{\text{test}} \): Subset used for testing the model.
\end{itemize}

The size of each subset is determined by:
\[
|D_{\text{test}}| = \text{test\_size} \times |D|, \quad |D_{\text{train}}| = |D| - |D_{\text{test}}|
\]
In this case, \(\text{test\_size} = 0.2\), so 20\% of the data is used for testing.

\subsection*{3) Tokenization}

Tokenization breaks text into smaller units called tokens (e.g., words or subwords). Each token is mapped to a numerical ID for processing. The tokenization process can be expressed as:
\[
X_{\text{tokens}} = \text{Tokenizer}(X_{\text{text}})
\]

\subsection*{Padding and Truncation}
To ensure uniform input lengths for the model:
\begin{itemize}
    \item \textbf{Padding}: Adds zeros to shorter sequences:
    \[
    X_{\text{padded}} = [\text{CLS}, x_1, x_2, \dots, x_n, \text{PAD}, \dots, \text{PAD}]
    \]
    \item \textbf{Truncation}: Trims longer sequences to the maximum length (128 in this case).
\end{itemize}

\subsection*{Numerical Input}
The numerical representation includes:
\begin{itemize}
    \item \textbf{Token IDs}: Numbers representing tokens.
    \item \textbf{Attention Masks}: Binary vectors indicating whether a token is meaningful (\(1\)) or padding (\(0\)).
\end{itemize}

\subsection*{4) Model Initialization}

DistilRoBERTa is a lighter version of the RoBERTa transformer. It uses the Transformer architecture, where attention mechanisms capture relationships between words:
\begin{equation}
Z = \text{softmax}\left(\frac{QK^T}{\sqrt{d_k}}\right)V
\end{equation}

Where:
\begin{itemize}
    \item \( Q \): Query matrix.
    \item \( K \): Key matrix.
    \item \( V \): Value matrix.
    \item \( d_k \): Dimensionality of keys.
\end{itemize}

The \([CLS]\) token's final embedding is passed to a classification head:
\begin{equation}
\hat{y} = W \cdot h_{\text{[CLS]}} + b
\end{equation}

Where:
\begin{itemize}
    \item \( W \): Weight matrix.
    \item \( h_{\text{[CLS]}} \): Embedding of the \([CLS]\) token.
    \item \( b \): Bias term.
\end{itemize}

\subsection*{5) Model Compilation}

\subsection*{Loss Function}
The model minimizes the sparse categorical cross-entropy loss, defined as:
\begin{equation}
L = - \frac{1}{N} \sum_{i=1}^{N} \sum_{j=1}^{C} y_{ij} \cdot \log(\hat{y}_{ij})
\end{equation}

Where:
\begin{itemize}
    \item \( N \): Number of samples.
    \item \( C \): Number of classes.
    \item \( y_{ij} \): Ground truth label for sample \(i\), class \(j\).
    \item \( \hat{y}_{ij} \): Predicted probability for sample \(i\), class \(j\).
\end{itemize}

\subsection*{Optimizer}
The Adam optimizer updates weights as:
\begin{equation}
W_t = W_{t-1} - \eta \cdot \frac{m_t}{\sqrt{v_t} + \epsilon}
\end{equation}

Where:
\begin{itemize}
    \item \( m_t \): Moving average of gradients.
    \item \( v_t \): Moving average of squared gradients.
    \item \( \eta \): Learning rate.
    \item \( \epsilon \): Smoothing term to prevent division by zero.
\end{itemize}

\subsection*{Learning Rate Schedule}
The learning rate decreases exponentially over time:
\begin{equation}
\eta_t = \eta_0 \cdot \gamma^{\lfloor t/\tau \rfloor}
\end{equation}

Where:
\begin{itemize}
    \item \( \eta_0 \): Initial learning rate.
    \item \( \gamma \): Decay rate (\(0.9\)).
    \item \( \tau \): Decay steps.
\end{itemize}

\subsection*{6) Training the Model}

The goal of training a machine learning model is to minimize the difference between the model's predictions (\( \hat{y} \)) and the true labels (\( y \)). This difference is quantified by a loss function.

\subsection*{Loss Function: Sparse Categorical Cross-Entropy}
For a multi-class classification problem, we use the Sparse Categorical Cross-Entropy Loss, defined as:
\begin{equation}
L = - \frac{1}{N} \sum_{i=1}^{N} \log P(y_i \mid X_i)
\end{equation}

Where:
\begin{itemize}
    \item \( N \): Number of samples.
    \item \( y_i \): True label for the \( i \)-th sample (as an integer index).
    \item \( P(y_i \mid X_i) \): Predicted probability for the true label \( y_i \), given input \( X_i \).
\end{itemize}

For a deep learning model like DistilRoBERTa, the probabilities are obtained using the softmax function:
\begin{equation}
P(y_i \mid X) = \frac{\exp(z_i)}{\sum_{j=1}^{C} \exp(z_j)}
\end{equation}

Where:
\begin{itemize}
    \item \( z_i \): Logit (raw score) for class \( i \).
    \item \( C \): Total number of classes.
\end{itemize}

\subsection*{7)Training Procedure}

\begin{itemize}
    \item \textbf{Forward Pass}: Compute model predictions (\( \hat{y} \)) for a mini-batch of training data.
    \item \textbf{Loss Computation}: Calculate the Sparse Categorical Cross-Entropy Loss.
    \item \textbf{Backward Pass}: Compute gradients of the loss function with respect to model parameters.
    \item \textbf{Parameter Update}: Update model parameters using the Adam optimizer.
\end{itemize}

\subsection*{Training Metrics}

During training, the following metrics are monitored:

\subsection*{Loss}
\begin{equation}
L_{\text{train}} = - \frac{1}{N_{\text{train}}} \sum_{i=1}^{N_{\text{train}}} \log P(y_i \mid X_i)
\end{equation}

\subsection*{Accuracy}
\begin{equation}
\text{Accuracy} = \frac{\text{Number of Correct Predictions}}{\text{Total Predictions}}
\end{equation}

\subsection*{Validation}
After each epoch, the model is evaluated on the validation set to monitor generalization. Validation metrics (\( L_{\text{val}}, \text{Accuracy}_{\text{val}} \)) are computed using the same formulas.

\subsection*{Model Evaluation}

After training, the model is evaluated on the test set.

\subsection*{Test Loss and Accuracy}
The test loss is calculated as:
\begin{equation}
L_{\text{test}} = - \frac{1}{N_{\text{test}}} \sum_{i=1}^{N_{\text{test}}} \log P(y_i \mid X_i)
\end{equation}

Test accuracy is calculated as:
\begin{equation}
\text{Accuracy}_{\text{test}} = \frac{\text{Number of Correct Predictions on Test Set}}{\text{Total Test Predictions}}
\end{equation}

DistilRoBERTa delivers strong sentiment analysis performance with its lightweight, efficient transformer architecture, balancing high accuracy and speed for large-scale or resource-constrained applications.\\

\section{Correlation}
This section explores the correlation between COVID-19 and monkeypox across three key dimensions: Healthcare System Trust, Economic Impact Perception, and Information Source Trust, highlighting the similarities and differences in public sentiment and societal responses to the two health crises.

\subsection{Correlation Analysis of Healthcare System Trust: COVID-19 vs. Monkeypox}
This study analyzes public trust in the healthcare system during the COVID-19 and Monkeypox crises using sentiment analysis, offering insights into variations in public perception and trust during these health emergencies.

\begin{figure}[htp]
    \centering
    \includegraphics[width=0.6\textwidth]{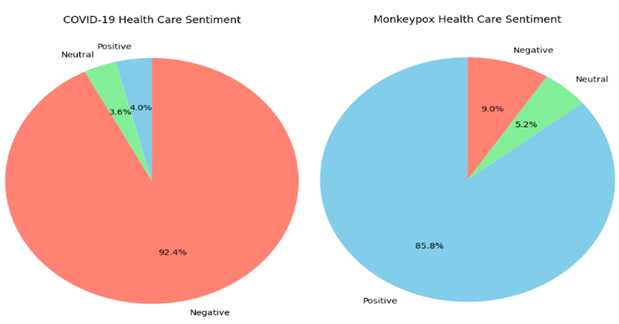}  
    \caption{Healthcare System Trust Comparison: COVID-19 vs. Monkeypox.}
    \label{fig:Trust}
\end{figure}

\subsection{Sentiment Analysis Result}

\begin{table}[h!]
    \centering
    \begin{tabular}{|l|c|c|}
        \hline
        \textbf{Sentiment} & \textbf{COVID-19 Sentiment (\%)} & \textbf{Monkeypox Sentiment (\%)} \\ 
        \hline
        Negative & 92.41 & 9.02 \\ 
        Positive & 3.99 & 85.81 \\ 
        Neutral  & 3.60 & 5.17 \\ 
        \hline
    \end{tabular}
    \caption{Sentiment distribution for public trust in healthcare systems.}
    \label{tab:helth}
\end{table}

Figure \ref{tab:helth} represents the sentiment distributions for public trust in healthcare systems during COVID-19 and Monkeypox:

\begin{itemize}
    \item \textbf{COVID-19 Healthcare Sentiment:} The sentiment is predominantly negative (92.4\%), with minimal positive (4.0\%) and neutral (3.6\%) sentiments.
    \item \textbf{Monkeypox Healthcare Sentiment:} While negative sentiment still dominates (85.8\%), the proportion of positive (5.2\%) and neutral (9.0\%) sentiments is slightly higher.
\end{itemize}

These results reveal a consistent lack of trust in the healthcare system during both crises, with variations in the intensity of public sentiment.

\subsection{Comparative Analysis}
The observed trends highlight significant similarities and differences:

\subsubsection{General Trends}
\begin{itemize}
    \item \textbf{COVID-19:} The overwhelming negative sentiment reflects widespread dissatisfaction with healthcare system responses to the pandemic, driven by global disruptions, perceived unpreparedness, and inconsistent communication. Trust in the healthcare system significantly correlates with willingness to test and vaccinate against COVID-19, with higher trust levels leading to increased willingness \cite{b26}.
    \item \textbf{Monkeypox:} The relatively lower negative sentiment suggests a less intense public response, potentially due to the localized and less severe nature of the outbreak. Healthcare providers demonstrated poorer knowledge regarding monkeypox compared to COVID-19, with only 45\% reporting adequate knowledge about monkeypox\cite{b27}
\end{itemize}

\subsubsection*{Differences in Public Sentiment}
\begin{itemize}
    \item \textbf{Severity and Scale:} COVID-19's global impact and healthcare strain amplified negative perceptions, while Monkeypox's localized nature led to less intense distrust.
    \item \textbf{Learning Effect:} The slight increase in positive and neutral sentiment for Monkeypox may reflect improved public perception of healthcare system management due to lessons learned from COVID-19.
\end{itemize}

\subsection{Statistical Insights}
To substantiate these findings quantitatively:

\begin{itemize}
    \item \textbf{Pearson Correlation:} Assess the relationship between sentiment proportions for both crises to establish the strength of correlation in public trust levels.
    \item \textbf{Chi-Square Test:} Evaluate whether the observed differences in sentiment distributions are statistically significant, highlighting the contrast in public responses.
\end{itemize}

These statistical analyses provide a robust foundation for interpreting sentiment variations.

\subsection{Interpretation of Public Perception}
Several factors may explain the observed differences in healthcare system trust during the two crises:

\begin{itemize}
    \item \textbf{Healthcare System Preparedness:} The unanticipated scale of COVID-19 led to public frustration, while the Monkeypox outbreak benefited from lessons learned and improved crisis management strategies.
    \item \textbf{Public Awareness and Communication:} Intense media scrutiny and inconsistent messaging during COVID-19 likely eroded trust, whereas Monkeypox elicited less media attention and public outcry.
    \item \textbf{Crisis Context:} The higher positive and neutral sentiments for Monkeypox may also reflect public perception of its relatively lower threat to individual health and society.
\end{itemize}

\subsection{Implications for Policy}
The findings underline the importance of maintaining and rebuilding public trust in healthcare systems during health crises:

\begin{itemize}
    \item \textbf{Strengthening Crisis Preparedness:} Proactive investment in healthcare infrastructure and planning is essential for mitigating public distrust.
    \item \textbf{Transparent Communication:} Clear, consistent, and evidence-based messaging can foster confidence in healthcare systems.
    \item \textbf{Learning from Past Crises:} The differences between COVID-19 and Monkeypox sentiment highlight the value of applying lessons from one crisis to improve responses to future emergencies.
\end{itemize}

By addressing public trust concerns, policymakers can ensure more resilient healthcare systems and more effective responses to future health crises.

\subsection{Correlation Analysis of Economic Sentiment in Healthcare Crises: COVID-19 vs. Monkeypox}
This study compares public sentiment on the economic impacts of COVID-19 and Monkeypox, using sentiment analysis to reveal varying perceptions of economic stability and financial implications during each crisis.\\
\pagebreak
\begin{figure}[htp]
    \centering
    \includegraphics[width= 14.5cm]{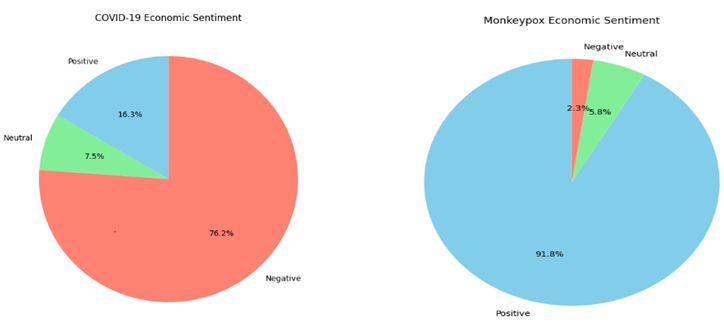}
    \caption{Economic Sentiment Comparison COVID vs Monkey pox}
    \label{fig:Trust}
\end{figure}
\newline
\subsection{Sentiment Analysis Result}

\begin{table}[h!]
\centering
\resizebox{\textwidth}{!}{%
\begin{tabular}{|l|c|c|}
\hline
\textbf{Sentiment} & \textbf{COVID Trust Sentiment (\%)} & \textbf{Monkeypox Trust Sentiment (\%)} \\ \hline
Positive & 7.5158 & 82.7945 \\ \hline
Negative & 86.9037 & 11.4319 \\ \hline
Neutral  & 3.83  & 7.53 \\ \hline
\end{tabular}%
}
\caption{Sentiment distribution for COVID and Monkeypox Economic Sentiment.}
\label{tab:econime}
\end{table}

Figure \ref{tab:econime} represents the sentiment distributions for the economic impacts of COVID-19 and Monkeypox:

\begin{itemize}
    \item \textbf{Monkeypox Economic Sentiment}: The public sentiment is predominantly positive (81.4\%), with neutral (9.1\%) and negative (9.5\%) sentiments forming smaller proportions.
    \item \textbf{COVID-19 Economic Sentiment}: Sentiment is overwhelmingly negative (86.9\%), with significantly lower levels of positive (5.6\%) and neutral (7.5\%) sentiment.
\end{itemize}

These results highlight substantial differences in public perception of the economic impacts of these two crises.

\subsection{Comparative Analysis}
The observed trends indicate notable differences in public sentiment:

\subsubsection{General Trend}
\begin{itemize}
    \item \textbf{Monkeypox}: The predominantly positive sentiment suggests an optimistic public perception of its limited economic impact.
    \item \textbf{COVID-19}: The dominance of negative sentiment reflects widespread financial distress caused by global economic disruptions, including lockdowns, unemployment, and economic instability.
\end{itemize}

\subsubsection{Differences in Impact}
\begin{itemize}
    \item \textbf{COVID-19} resulted in unprecedented global economic challenges, leading to severe public dissatisfaction.
    \item \textbf{Monkeypox}, with its more localized and less severe economic implications, likely evoked less economic concern.
\end{itemize}

These differences illustrate how the scope and severity of a health crisis influence public perceptions of economic stability.

\subsection{Statistical Insights}
To support these observations quantitatively:

\begin{itemize}
    \item \textbf{Pearson Correlation Analysis}: Assess the strength of the relationship between sentiment proportions across the two crises.
    \item \textbf{Chi-square Test}: Determine whether the observed differences in sentiment distribution are statistically significant.
\end{itemize}

The statistical insights reinforce the comparative analysis, highlighting the distinct public perceptions of the economic impacts of COVID-19 and Monkeypox.

\subsection{Interpretation of Public Perception}
Several factors may have contributed to the observed differences in economic sentiment:

\begin{itemize}
    \item \textbf{Scale of Economic Disruption}: COVID-19's global impact and prolonged economic challenges starkly contrast with Monkeypox's localized and less severe economic effects.
    \item \textbf{Media Coverage and Public Awareness}: The intense focus on COVID-19's economic ramifications likely amplified negative public sentiment, while Monkeypox received comparatively limited attention.
    \item \textbf{Crisis Preparedness}: The public perception of economic management during Monkeypox may have been influenced by lessons learned from COVID-19.
\end{itemize}

\subsection{Implications for Policy}
The findings provide valuable insights for policymakers:

\begin{itemize}
    \item \textbf{Proactive Economic Interventions}: Governments must implement timely measures to minimize economic disruption during health crises.
    \item \textbf{Public Communication}: Clear and transparent messaging about economic policies can help alleviate public concerns and foster trust.
    \item \textbf{Resilience Planning}: Lessons from the COVID-19 and Monkeypox crises can inform strategies to build economic resilience for future public health emergencies.
\end{itemize}

By addressing public sentiment during health crises, governments can better prepare for and mitigate the economic fallout of future challenges.

\subsection{Correlation Analysis of Information Source Trust in Healthcare Crises: COVID-19 vs. Monkeypox}
This study compares public trust in information sources during the COVID-19 and Monkeypox crises, using sentiment analysis to assess variations in the perception of reliability and trustworthiness across these health emergencies.

\pagebreak
\begin{figure}[htp]
    \centering
    \includegraphics[width= 10.5cm]{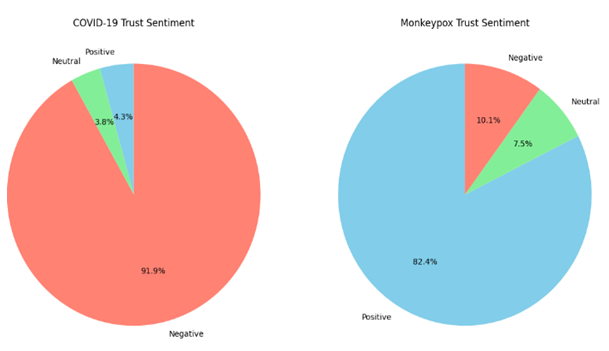}
    \caption{Information Source Trust  Comparison COVID vs Monkey pox}
    \label{fig:info}
\end{figure}

\subsection{Sentiment Analysis Result}

\begin{table}[h!]
\centering
\resizebox{\textwidth}{!}{%
\begin{tabular}{|l|c|c|}
\hline
\textbf{Sentiment} & \textbf{COVID Trust Sentiment (\%)} & \textbf{Monkeypox Trust Sentiment (\%)} \\ \hline
Positive & 4.3027 & 82.4002 \\ \hline
Negative & 91.8694 & 10.0723 \\ \hline
Neutral  & 3.8279 & 7.5275 \\ \hline
\end{tabular}%
}
\end{table}

Figure \ref{fig:info} represent the sentiment distributions reflecting public trust in information sources during the economic crises of COVID-19 and Monkeypox:

\begin{itemize}
    \item \textbf{COVID-19 Information Source Trust}: Sentiment is overwhelmingly negative (91.9\%), with significantly lower levels of positive (4.3\%) and neutral (3.8\%) sentiment.
    \item \textbf{Monkeypox Information Source Trust}: Public sentiment is predominantly positive (82.4\%), with neutral (7.5\%) and negative (10.1\%) sentiments forming smaller proportions.
\end{itemize}

These results highlight striking differences in public trust toward information sources between the two crises.

\subsection{Comparative Analysis}

The observed trends indicate significant variations in public sentiment:

\subsubsection{General Trend}

\begin{itemize}
    \item \textbf{COVID-19:} The overwhelming negative sentiment (91.9\%) indicates a significant erosion of trust in information sources, likely fueled by misinformation, conflicting statements, and perceived mishandling during the pandemic.
    \item \textbf{Monkeypox:} The high percentage of positive sentiment (82.4\%) suggests strong public confidence in the information sources during this crisis.
    
\end{itemize}

\subsubsection{Differences in Trust}

\begin{itemize}
    \item \textbf{COVID-19:} The high negative sentiment likely stems from the overwhelming volume of contradictory information, the early stages of the crisis, and confusion surrounding public health directives and responses.
    \item \textbf{Monkeypox:} The predominantly positive sentiment could reflect the more localized and contained nature of the outbreak, with a clear flow of information that did not face the widespread misinformation issues seen with COVID-19.
    
\end{itemize}

These differences highlight how the scale, duration, and information environment of a health crisis influence the public's trust in information sources.

\subsection{Statistical Insights}

To support these observations quantitatively:

\begin{itemize}
    \item \textbf{Pearson Correlation:} A correlation analysis can reveal how trust sentiment across the two crises relates, with a negative correlation indicating that higher crisis intensity and media coverage lead to more negative trust.
    \item \textbf{Chi-Square Test:} This statistical test can determine whether the differences in information source trust distributions between the two crises are statistically significant.
\end{itemize}

These insights offer a deeper understanding of how the nature of each crisis influences public trust in the information provided by governments, media, and health authorities.

\subsection{Interpretation of Public Perception}

Several factors may explain the observed differences in trust:

\begin{itemize}
    \item \textbf{Crisis Scope and Duration:} COVID-19's prolonged global impact likely increased skepticism towards information sources, while Monkeypox's localized nature may have fostered clearer, more consistent communication.
    \item \textbf{Misinformation and Communication Breakdown:} The volume of misinformation during COVID-19 eroded trust, whereas Monkeypox's more contained nature likely led to clearer, more reliable information.
    \item \textbf{Public Health Messaging:} COVID-19 faced challenges with shifting guidelines and uncertainty in government communications, while more controlled messaging during Monkeypox likely fostered higher trust levels.
\end{itemize}

\subsection{Implications for Policy}

The findings highlight key lessons for policymakers:

\begin{itemize}
    \item \textbf{Improved Communication Strategy:} A clear, consistent, and transparent communication strategy is crucial in maintaining public trust during a health crisis.
    \item \textbf{Combatting Misinformation:} Proactive measures to combat misinformation are essential to prevent public skepticism and confusion.
    \item \textbf{Crisis-Specific Approaches:} Understanding the unique dynamics of each crisis can help tailor communication strategies that resonate with the public and reinforce trust.
\end{itemize}

This comparative analysis of information source trust between COVID-19 and Monkeypox provides insights into the impact of crisis management and communication on public sentiment, serving as a guide for future crisis communication efforts.

\section{Result Analysis \& Discussion}
This section outlines the experimental steps used to evaluate five models based on labeling, vectorization, and normalization methods. Two data labeling approaches were used: the first with Logistic Regression and Naïve Bayes, and the second with three transformer-based models, with performance measured by accuracy, precision, recall, and F1 score.
\subsection{Algorithms comparison on approach 1}

\begin{table}[ht]
    \centering
    \caption{Algorithms comparison: COVID-19}
    \vspace{0.5em}
    \label{tab:algorithm_comparison_covid}
    \begin{tabular}{|p{3.5cm}|c|c|c|c|}
        \hline
        \textbf{Model} & \textbf{Accuracy} & \textbf{Precision} & \textbf{Recall} & \textbf{F1 Score} \\
        \hline
        Logistic Regression & 87.69\% & 0.88 & 0.62 & 0.62 \\ \hline
        Naïve Bayes & 70.00\% & 0.89 & 0.40 & 0.44 \\ 
        \hline
    \end{tabular}
\end{table}

\vspace{1em}

\begin{table}[ht]
    \centering
    \caption{Algorithms comparison: Monkeypox}
    \vspace{0.5em}
    \label{tab:algorithm_comparison_monkeypox}
    \begin{tabular}{|p{3.5cm}|c|c|c|c|}
        \hline
        \textbf{Model} & \textbf{Accuracy} & \textbf{Precision} & \textbf{Recall} & \textbf{F1 Score} \\
        \hline
        Logistic Regression & 82.96\% & 0.83 & 0.52 & 0.56 \\ \hline
        Naïve Bayes & 89.69\% & 0.79 & 0.41 & 0.44 \\ 
        \hline
    \end{tabular}
\end{table}

\subsection{Comprehensive Sentiment Analysis}
This study involved sentiment analysis, where the dataset was preprocessed and labeled using a pre-trained RoBERTa model for robust sentiment classification, followed by the application of Logistic Regression and Naïve Bayes to evaluate performance.

\subsection{Model Performance Comparison}

The evaluation of Logistic Regression and Naïve Bayes on the labeled data revealed distinct differences in their ability to classify sentiments:

\begin{itemize}
    \item \textbf{Logistic Regression} achieved superior performance with higher accuracy (87.69\%) and F1 Score (0.62). This indicates its strong ability to balance precision and recall, making it well-suited for tasks where overall classification performance is critical.
    \item \textbf{Naïve Bayes} achivimg 70\% accuracy with achieving a higher precision (89.00) compared to Logistic Regression (88.00), struggled with recall (0.40) and, consequently, had a lower F1 Score (0.44).
\end{itemize}

For the labeled data related to Monkeypox sentiments, the models exhibited a similar trend:

\begin{itemize}
    \item \textbf{Logistic Regression} demonstrated an accuracy of 82.96\% and an F1 Score of 0.56, outperforming Naïve Bayes, which recorded an accuracy of 89.69\% and an F1 Score of 0.44.
    \item \textbf{Naïve Bayes} again showed higher precision (89.69\%) but a significantly lower recall (0.41), reinforcing its limitation in identifying the full range of sentiment categories.
\end{itemize}

\subsection{Insights from Sentiment Analysis}

The results highlight key observations:
\begin{enumerate}
    \item \textbf{Model Robustness:} Logistic Regression consistently outperformed Naïve Bayes, demonstrating its robustness in handling the labeled data. Its ability to balance precision and recall makes it a reliable model for sentiment classification tasks.
    \item \textbf{Trade-offs in Precision and Recall:} Naïve Bayes exhibited a tendency towards higher precision at the expense of recall. While this might be suitable for tasks requiring minimal false positives, its overall performance (as indicated by the F1 Score) was suboptimal compared to Logistic Regression.
    \item \textbf{Consistency Across Datasets:} The observed trends were consistent across sentiments related to COVID-19 and Monkeypox, suggesting the generalizability of the findings.
\end{enumerate}

\section{Implications and Future Directions}

The findings have significant implications for sentiment analysis in the context of health crises:

\begin{itemize}
    \item \textbf{Model Selection:} Logistic Regression is recommended for tasks requiring balanced sentiment classification, while Naïve Bayes may be used in scenarios prioritizing precision, provided its recall limitations are addressed.
    \item \textbf{Labeling with Pre-trained Models:} The use of RoBERTa for data labeling provided a robust foundation for the sentiment analysis, highlighting the value of integrating pre-trained models into traditional workflows.
    \item \textbf{Future Improvements:} Incorporating ensemble models or feature engineering techniques may further enhance the classification performance, particularly for datasets with unique characteristics, such as those related to emerging health crises.
\end{itemize}
This study combines advanced labeling techniques with traditional machine learning models to provide comprehensive insights into public trust and healthcare sentiment evaluation.\\

\subsection{Algorithm Comparison on Approach 2}

\begin{table}[ht]
    \centering
    \caption{Algorithms Comparison: COVID-19}
    \vspace{0.5em}
    \label{tab:algorithm_comparison_approach2_covid}
    \begin{tabular}{|p{3.5cm}|c|c|c|c|}
        \hline
        \textbf{Model} & \textbf{Accuracy (\%)} & \textbf{Precision} & \textbf{Recall} & \textbf{F1 Score} \\
        \hline
        RoBERTa & 97.99 & 0.95 & 0.95 & 0.95 \\ \hline
        XLNet & 97.05 & 0.97 & 0.97 & 0.97 \\ \hline
        DistilRoBERTa & 97.51 & 0.97 & 0.97 & 0.98 \\
        \hline
    \end{tabular}
\end{table}

\vspace{1em}

\begin{table}[ht]
    \centering
    \caption{Algorithms Comparison: Monkeypox}
    \vspace{0.5em}
    \label{tab:algorithm_comparison_approach2_monkeypox}
    \begin{tabular}{|p{3.5cm}|c|c|c|c|}
        \hline
        \textbf{Model} & \textbf{Accuracy (\%)} & \textbf{Precision} & \textbf{Recall} & \textbf{F1 Score} \\
        \hline
        RoBERTa & 95.36 & 0.95 & 0.95 & 0.95 \\ \hline
        XLNet & 96.82 & 0.97 & 0.97 & 0.97 \\ \hline
        DistilRoBERTa & 95.19 & 0.95 & 0.95 & 0.95 \\
        \hline
    \end{tabular}
\end{table}

\subsection{Comprehensive Sentiment Analysis}
This study involved a comprehensive sentiment analysis. The dataset was initially preprocessed and labeled using TextBlob, which uses a combination of rule-based and lexicon-based methods to classify sentiments into positive, neutral, and negative categories. Subsequently, Transformer-based models RoBERTa , XLNet and DistilRoberta were applied to the labeled data to evaluate their performance in sentiment classification.

\subsection{Model Performance Comparison}

The evaluation of RoBERTa, XLNet, and DistilRoBERTa on the sentiment analysis datasets revealed notable differences in their classification performance:

\begin{itemize}
    
\item\textbf{RoBERTa:}  
RoBERTa achieved strong overall performance for both COVID-19 and Monkeypox datasets. For the COVID-19 dataset, it recorded the highest accuracy (\textbf{97.99\%}) and balanced Precision, Recall, and F1 Score (all \textbf{0.95}), indicating its robustness in classifying sentiments effectively. On the Monkeypox dataset, RoBERTa maintained a high accuracy of \textbf{95.36\%} with equally balanced metrics (\textbf{0.95}). Its performance demonstrates its reliability in consistently capturing sentiment patterns across datasets.

\item\textbf{XLNet:}  
XLNet demonstrated remarkable performance, particularly on the Monkeypox dataset, where it achieved the highest accuracy of \textbf{96.82\%} along with Precision, Recall, and F1 Score values of \textbf{0.97}. For the COVID-19 dataset, XLNet displayed slightly lower accuracy (\textbf{97.05\%}) but matched DistilRoBERTa with high Precision, Recall, and F1 Score (\textbf{0.97}), highlighting its strength in minimizing false positives and false negatives.

\item\textbf{DistilRoBERTa:}  
DistilRoBERTa showed a unique balance between accuracy and F1 Score. On the COVID-19 dataset, it recorded an accuracy of \textbf{97.51\%} and achieved the highest F1 Score (\textbf{0.98}), showcasing its superior ability to balance Precision and Recall. However, on the Monkeypox dataset, DistilRoBERTa performed similarly to RoBERTa, with an accuracy of \textbf{95.19\%} and consistent metrics (\textbf{0.95}). This indicates its versatility across datasets while maintaining a strong balance in performance.

\end{itemize}

\subsection{Insights from Sentiment Analysis}
The evaluation of RoBERTa, XLNet, and DistilRoBERTa revealed several key observations:
\begin{enumerate}
    \item \textbf{Model Robustness:} All three models demonstrated robust performance across the COVID-19 and Monkeypox datasets. XLNet consistently delivered the highest accuracy on the Monkeypox dataset (\textbf{96.82\%}) and maintained strong metrics across both datasets, indicating its adaptability and reliability in sentiment classification tasks. Similarly, DistilRoBERTa showcased superior performance on the COVID-19 dataset, achieving the highest F1 Score (\textbf{0.98}), reflecting its ability to balance Precision and Recall effectively.
    \item \textbf{Trade-offs in Metrics:} While RoBERTa provided balanced Precision, Recall, and F1 Scores across datasets, XLNet exhibited slight metric trade-offs depending on the dataset, excelling in Accuracy and F1 Score for Monkeypox but slightly lagging in COVID-19 performance. DistilRoBERTa's consistently high F1 Score indicates its suitability for tasks requiring an optimal balance of Precision and Recall.
    \item \textbf{Consistency Across Datasets:} The observed trends in model performance were consistent across both COVID-19 and Monkeypox datasets. RoBERTa demonstrated steady and reliable performance in both cases, underscoring its generalizability. Meanwhile, the dataset-specific strengths of XLNet and DistilRoBERTa suggest that the choice of model may depend on the characteristics of the dataset and the specific requirements of the sentiment classification task.
\end{enumerate}

\subsection{Implications and Future Directions}
The findings of this study have significant implications for sentiment analysis in the context of health crises, particularly with the evaluation of transformer-based models like RoBERTa, XLNet, and DistilRoBERTa:
\begin{itemize}
    \item \textbf{Model Selection:} For tasks requiring balanced sentiment classification,\\ RoBERTa is a reliable choice due to its consistent performance across both COVID-19 and Monkeypox datasets. XLNet is particularly effective for datasets where high accuracy is prioritized, as demonstrated by its superior results on the Monkeypox dataset. DistilRoBERTa, with its exceptional F1 Score on the COVID-19 dataset, is recommended for tasks emphasizing a balance between Precision and Recall.
    \item \textbf{Leveraging Pre-trained Models:} The strong performance of all three models highlights the importance of pre-trained transformer architectures in sentiment analysis. These models excel in handling complex datasets, emphasizing their value in scenarios where domain-specific labeled data may be limited.
    \item \textbf{Future Improvements:} Incorporating ensemble techniques that combine the strengths of RoBERTa, XLNet, and DistilRoBERTa could further enhance classification performance. Additionally, applying advanced feature engineering or fine-tuning these models on domain-specific datasets could improve their adaptability to unique challenges, such as sentiment analysis during emerging health crises.
\end{itemize}

By utilizing advanced transformer-based models, This study employs advanced transformer-based models to deliver robust sentiment analysis, offering key insights into public concerns and informing future healthcare sentiment evaluation.

\section{Limitations}
This study faced limitations including challenges in capturing nuanced health-related sentiments, a restricted range of algorithms, and the need for more diverse datasets, suggesting future improvements through broader data and advanced modeling techniques.\\

\section{Future work}
This research plans to expand by using the Twitter API for real-time sentiment analysis, enabling continuous monitoring of public reactions during health crises.
Future work includes applying advanced techniques to correlate sentiment trends with key health events for deeper insights.
Improvements in data preprocessing and domain-specific feature engineering are also targeted to boost model accuracy.
These enhancements aim to support policymakers with timely, data-driven insights for effective public health responses.\\

\section{Conclusion}
This study compared sentiment analysis models on COVID-19 and Monkeypox data, showing transformer models like RoBERTa outperform traditional ones.
It highlighted the importance of model selection and the strength of pre-trained architectures in capturing sentiment accurately.
Future work includes real-time data integration, refined feature engineering, and analyzing correlations to support better public health decision-making.

\section*{Acknowledgments}

\bibliographystyle{unsrt}  
\bibliography{references}

\begin{thebibliography}{}

\end{thebibliography}


\begin{thebibliography}{99}

\bibitem{b1}
Melton C, White B, Davis R, Bednarczyk R, Shaban-Nejad A.  
Fine-tuned Sentiment Analysis of COVID-19 Vaccine–Related Social Media Data: Comparative Study.  
\textit{J Med Internet Res}. 2022;24(10):e40408.  
Available: \url{https://www.jmir.org/2022/10/e40408}  
DOI: \href{https://doi.org/10.2196/40408}{10.2196/40408}

\bibitem{b2}
Staphord Bengesi, Timothy Oladunni, Ruth Olusegun, et al.  
A Machine Learning-Sentiment Analysis on Monkeypox Outbreak: An Extensive Dataset to Show the Polarity of Public Opinion From Twitter Tweets.  
\textit{IEEE Access}. 2023;11:11811-11826.  
DOI: \href{https://doi.org/10.1109/ACCESS.2023.3242290}{10.1109/ACCESS.2023.3242290}

\bibitem{b3}
Al-Ahdal T, Akbar M, Khan S, et al.  
Improving Public Health Policy by Comparing the Public Response during the Start of COVID-19 and Monkeypox on Twitter in Germany: A Mixed Methods Study.  
\textit{Vaccines}. 2022;10(12):1985.  
DOI: \href{https://doi.org/10.3390/vaccines10121985}{10.3390/vaccines10121985}

\bibitem{b4}
Thakur N.  
Sentiment Analysis and Text Analysis of the Public Discourse on Twitter about COVID-19 and MPox.  
\textit{Big Data and Cognitive Computing}. 2023;7(2):116.  
DOI: \href{https://doi.org/10.3390/bdcc7020116}{10.3390/bdcc7020116}

\bibitem{b5}
BBC News.  
First case of more dangerous mpox found outside Africa.  
\textit{BBC News}. 2024, August 16.  
Available: \url{https://www.bbc.com/news/articles/c4gqr5lrpwxo}

\bibitem{b6}
World Health Organization.  
Episode \#76 - Monkeypox: Who is at risk? [Podcast episode].  
\textit{World Health Organization}. 2022, July 23.  
Available: \url{https://www.who.int/podcasts/episode/science-in-5/episode--76---monkeypox--who-is-at-risk}

\bibitem{b7}
Bengesi S, Oladunni T, Olusegun R, Audu H.  
A Machine Learning-Sentiment Analysis on Monkeypox Outbreak: An Extensive Dataset to Show the Polarity of Public Opinion From Twitter Tweets.  
\textit{IEEE Access}. 2023;11:11811-11826.  
DOI: \href{https://doi.org/10.1109/ACCESS.2023.3242290}{10.1109/ACCESS.2023.3242290}  
Keywords: Sentiment analysis, monkeypox, Twitter, machine learning, TF-IDF, TextBlob, VADER.

\bibitem{b8}
Manning CD, Raghavan P, Schütze H.  
\textit{Introduction to Information Retrieval}.  
Cambridge: Cambridge University Press; 2008.

\bibitem{b9}
Dileep Kumar M, Soumya Ranjan Jena.  
\textit{NLP for Sentiment Analysis}.  
Xoffencer International Book Publication House, 2024.  
ISBN: 9789348116611.  
DOI: \href{https://doi.org/10.5281/zenodo.14178332}{10.5281/zenodo.14178332}  
x RESEARCH \& INNOVATION LAB.

\bibitem{b10}
Majid A, Ahmed M, Khan M, et al.  
Sentiment Analysis on Tiktok Application Reviews Using Natural Language Processing Approach.  
\textit{Journal of Embedded Systems, Security and Intelligent Systems}. 2023.

\bibitem{b11}
Jianqiang Z, Xiaolin G.  
Comparison Research on Text Pre-processing Methods on Twitter Sentiment Analysis.  
\textit{IEEE Access}. 2017;5:2870-2879.  
DOI: \href{https://doi.org/10.1109/ACCESS.2017.2672677}{10.1109/ACCESS.2017.2672677}  
Keywords: Twitter, Sentiment analysis, Terminology, Support vector machines, Uniform resource locators, Organizations, Analytical models.

\bibitem{b12}
Singh G, Vikram G, et al.  
Predicting multi-label emojis, emotions, and sentiments in code-mixed texts using an emojifying sentiments framework.  
\textit{Scientific Reports}. 2024;14(1):12204.  
DOI: \href{https://doi.org/10.1038/s41598-024-58944-5}{10.1038/s41598-024-58944-5}

\bibitem{b13}
Liu Y, et al.  
RoBERTa: A Robustly Optimized BERT Pretraining Approach.  
\textit{ArXiv}. 2019.  
Available: \url{https://arxiv.org/abs/1907.11692}

\bibitem{b15}
Suanpang P, Jamjuntr P, Kaewyong P.  
Sentiment Analysis with a TextBlob Package Implications for Tourism.  
\textit{Research Article}. 2021;24(6S).  
Published by Suan Dusit University and King Mongkut's University of Technology Thonburi.

\bibitem{b16}
Bengesi S, Oladunni T, Olusegun R, Audu H.  
A Machine Learning-Sentiment Analysis on Monkeypox Outbreak: An Extensive Dataset to Show the Polarity of Public Opinion From Twitter Tweets.  
\textit{IEEE Access}. 2023;11:11811-11826.  
DOI: \href{https://doi.org/10.1109/ACCESS.2023.3242290}{10.1109/ACCESS.2023.3242290}  
Keywords: Social networking, Sentiment analysis, Blogs, Classification algorithms, Machine learning, Monkeypox, TF-IDF, TextBlob, VADER.

\bibitem{b17}
Wang Y, et al.  
Comparisons and Selections of Features and Classifiers for Short Text Classification.  
\textit{IOP Conference Series: Materials Science and Engineering}. 2017;261.  
Available: \url{https://iopscience.iop.org/article/10.1088/1757-899X/261/1/012050}

\bibitem{b18}
Liu Y, Ott M, Goyal N, Du J, Joshi M, Chen D, et al.  
RoBERTa: A robustly optimized BERT pretraining approach.  
\textit{arXiv}. 2019.  
Available: \url{https://arxiv.org/abs/1907.11692}

\bibitem{b19}
The General Language Understanding Evaluation (GLUE) Benchmark.  
Available: \url{https://gluebenchmark.com/leaderboard}, Sep. 2021.

\bibitem{b20}
Rajapaksha P, Farahbakhsh R, Crespi N.  
BERT, XLNet or RoBERTa: The Best Transfer Learning Model to Detect Clickbaits.  
\textit{IEEE Access}. 2021;9:154704-154716.  
DOI: \href{https://doi.org/10.1109/ACCESS.2021.3128742}{10.1109/ACCESS.2021.3128742}  
Keywords: Transfer learning, Clickbait, Fake news, BERT, RoBERTa, XLNet, Twitter.

\bibitem{b21}
Janosch H, Poesio M.  
Patterns of Polysemy and Homonymy in Contextualised Language Models.  
\textit{2021}.  
Pages 2663-2676.

\bibitem{b22}
Agarwala A, Schoenholz S, Pennington J, Dauphin Y.  
Temperature check: theory and practice for training models with softmax-cross-entropy losses.  
\textit{arXiv}. 2021.  
Available: \url{https://arxiv.org/abs/2102.08763}

\bibitem{b23}
Mohamed S, Shah S, Abuaieta MA, Saeed S, Almazrouei S.  
Safeguarding Online Communications using DistilRoBERTa for Detection of Terrorism and Offensive Chats.  
\textit{Journal of Information Security and Cybercrimes Research}. 2024.  
DOI: \href{https://doi.org/10.26735/vnvr2791}{10.26735/vnvr2791}

\bibitem{b24}
Raj K, Goswami B, Mhatre SM, Agrawal S.  
Naive Bayes in Focus: A Thorough Examination of its Algorithmic Foundations and Use Cases.  
\textit{International Journal of Innovative Science and Research Technology}. 2024.  
DOI: \href{https://doi.org/10.38124/ijisrt/ijisrt24may1438}{10.38124/ijisrt/ijisrt24may1438}

\bibitem{b25}
Yaddanapudi L, Hahn J, Ladikas M.  
A multifaceted analysis of decreasing trust in health institutions in the EU during the COVID-19 pandemic.  
\textit{Deleted Journal}. 2024.  
DOI: \href{https://doi.org/10.1186/s12982-024-00240-8}{10.1186/s12982-024-00240-8}

\bibitem{b26}
Taiwo, Oluwaseun, Sokunbi., J., Omojuyigbe. "4. Re-Emergence of Monkeypox Amidst COVID-19 Pandemic in Africa: What is the Fate of the African Healthcare System?."  undefined (2023). doi: 10.36108/gjoboh/3202.20.0140

\bibitem{b26}
Judy, Nanaw., Juliana, S., Sherchan., Jessica, R., Fernandez., Paula, D, Strassle., Wizdom, Powell., Allana, T., Forde. "Racial/ethnic differences in the associations between trust in the U.S. healthcare system and willingness to test for and vaccinate against COVID-19." BMC Public Health, 24 (2024). doi: 10.1186/s12889-024-18526-6

\bibitem{b27}
Sarya, Swed., Hidar, Alibrahim., H., Bohsas., Nagham, Jawish., M., A., Rais., Mohamad, Nour, Nasif., Wael, Hafez., Bisher, Sawaf., Ahmed, H., Abdelrahman., Sherihan, Fathey., I., Atef, Ismail, Ahmed, Ibrahim., A., Rakab., Mohamed, Elsayed. "4. A multinational cross-sectional study on the awareness and concerns of healthcare providers toward monkeypox and the promotion of the monkeypox vaccination." Frontiers in Public Health, undefined (2023). doi: 10.3389/fpubh.2023.1153136

\bibitem{b28}
Kayal, Padmanandam., Sai, Priya, V, D, S, Bheri., LaxmiHarshika, Vegesna., Kalakuntla, Sruthi. (2021). A Speech Recognized Dynamic Word Cloud Visualization for Text Summarization.   doi: 10.1109/ICICT50816.2021.9358693

\bibitem{b29}
Kang, Feng., Alice, Gao., Johanna, Suvi, Karras. (2022). 2. Towards Semantically Aware Word Cloud Shape Generation.   doi: 10.1145/3526114.3558724

\bibitem{b30}
Morteza, Abdullatif, Khafaie., Fakher, Rahim. "6. Cross-country comparison of case fatality rates of Covid-19/SARS-CoV-2." Osong public health and research perspectives, undefined (2020). doi: 10.24171/J.PHRP.2020.11.2.03

\bibitem{b31}
Dominic, Cortis. (2020). 1. On Determining the Age Distribution of COVID-19 Pandemic. Frontiers in Public Health,  doi: 10.3389/FPUBH.2020.00202

\bibitem{b32}
Tarun, Kumar, Suvvari., Mokanpally, Sandeep., Jogender, Kumar., Prakasini, Satapathy., Santenna, Chenchula., Aravind, P, Gandhi., Muhammad, Aaqib, Shamim., Patricia, Schlagenhauf., Alfanso, J, Rodriguez-Morales., Ranjit, Sah., Keerti, Bhusan, Pradhan., Sarvesh, Rustagi., Alaa, H, Hermis., Bijaya, Kumar, Padhi. "1. A meta‐analysis and mapping of global mpox infection among children and adolescents." Reviews in Medical Virology, undefined (2023). doi: 10.1002/rmv.2472

\bibitem{b33}
Patrick, Eustaquio., LaTweika, A, T, Salmon-Trejo., Lisa, C., McGuire., Sascha, R., Ellington. "1. Epidemiologic and Clinical Features of Mpox in Adults Aged >50 Years — United States, May 2022–May 2023."  undefined (2023). doi: 10.15585/mmwr.mm7233a3

\bibitem{b34}
Navnita, Kisku., Tushar, Agarwal., Dikshant, Jain. "2. COVID-19 Impact on Female Patients at Tertiary Care Hospital – A Retrospective Study."  undefined (2024). doi:10.25259/ijcdw\_50\_2023

\bibitem{b35}
Carolina, Coutinho., Mayara, Secco, Torres, Silva., Thiago, S., Torres., Eduardo, Peixoto., Monica, Avelar, Magalhães., Sandra, Wagner, Cardoso., Gabriela, Nazário., Maíra, Mendonça., Mariana, Menezes., Paula, Maria, Almeida., Paula, Rita, Dias, de, Brito, de, Carvalho., Shenon, Bia, Bedin., Aline, Maria, Almeida., Silvia, Carvalho., Valdilea, Gonçalves, Veloso., Beatriz, Grinsztejn., Luciane, de, Souza, Velasque., André, M., Japiassú., Marcel, Trepow., Italo, Guariz, Ferreira., Larissa, Villela., Rafael, Teixeira, Fraga., Mariah, Castro, de, Souza, Pires., R., O., da, Silva, Escada., Leonardo, Paiva, de, Sousa., Gabriela, Lisseth, Umaña, Robleda., D., V., Santos., Luiz, Ricardo, Siqueira, Camacho., Pedro, Amparo., João, Victor, Jaegger, de, França., Felipe, de, Oliveira, Heluy, Correa., Bruno, Ivanovinsky, Costa, de, Sousa., Bernardo, Vicari, do, Valle., João, Paulo, Bortot, Soares., Livia, Cristina, Fonseca, Ferreira., P., da, Silva, Martins., M., B., Mesquita., José, Ricardo, Coutinho., Raissa, de, Moraes, Perlingeiro., Priscila, Peixoto, de, Castro, Oliveira., Hugo, Perazzo, Pedroso, Barbosa., André, Figueiredo, Accetta., M., Cunha., R., Eiras., Ticiana, Martins, dos, Santos., Wladmyr, Davila, da, Silva., Monique, do, Vale, da, Silveira., Tania, de, Souza, Brum., Guilherme, Amaral, Calvet., Rodrigo, Caldas, Menezes., Sandro, Antônio, Pereira. "2. Characteristics of women diagnosed with mpox infection compared to men: A case series from Brazil.." Travel Medicine and Infectious Disease, undefined (2023). doi: 10.1016/j.tmaid.2023.102663

\bibitem{b36}
CDC, "Mpox in the U.S.," \textit{Centers for Disease Control and Prevention}, Jul. 22, 2022. [Online]. Available: \url{https://www.cdc.gov/poxvirus/monkeypox/about/index.html}. [Accessed: Dec. 23, 2022].

\bibitem{b37}
 A. Mandavilli, "W.H.O. Declares Monkeypox Spread a 
Global Health Emergency," The New York Times, Jul. 23, 
2022. Accessed: Dec. 23, 2022. [Online]. Available: 
https://www.nytimes.com/2022/07/23/health/monkeypox
pandemic-who.html 

\bibitem{b38}
 A. S. for P. Affairs (ASPA), "Biden-Harris 
Administration Bolsters Monkeypox Response; HHS 
Secretary Becerra Declares Public Health Emergency,", 
HHS.gov,
 Aug.
 04,
 2022. 
https://www.hhs.gov/about/news/2022/08/04/biden-harris
administration-bolsters-monkeypox-response-hhs
secretary-becerra-declares-public-health-emergency.html 
(accessed Dec. 23, 2022). 

\bibitem{b39}
 J.-H. Yoo et al., "Once Bitten, Twice Shy: Our Attitude 
Towards Monkeypox," J. Korean Med. Sci., vol. 37, no. 22, 
p. e188, 2022. 

\bibitem{b40}
CDC, "2022 U.S. Map \& Case Count," Dec. 22, 2022. 
https://www.cdc.gov/poxvirus/monkeypox/response/2022/u
 s-map.html (accessed Dec. 23, 2022). 

\bibitem{b41}
 Hutchinson, A. New Study Shows Twitter Is the Most Used Social Media Platform among Journalists. Social Media Today, 28 June 2022. Available online: https://www.socialmediatoday.com/news/new-study-shows-twitter-is-the-most-used-social media-platform-among-journa/626245/ (accessed on 26 March 2023).
 
\bibitem{b42}
 U. Naseem, I. Razzak, M. Khushi, P. W. Eklund, and J. 
Kim, "COVIDSenti: A large-scale benchmark Twitter data 
set for COVID-19 sentiment analysis," IEEE Trans. Comput. 
Soc. Syst., vol. 8, no. 4, pp. 1003–1015, 2021. 

\bibitem{b43}
F. Shamrat, M. S. Chakrabarty, S. K. Das, S. Hossain, and J. S. Alam, 
"Sentiment analysis on Twitter tweets about COVID-19 vaccines using NLP and supervised KNN classification algorithm," 
\textit{Indonesian Journal of Electrical Engineering and Computer Science}, 
vol. 23, no. 1, pp. 463--470, 2021. 
doi: 10.11591/ijeecs.v23.i1.pp463-470. \% Add DOI if applicable.

\end{thebibliography}

\end{document}